\documentclass{sigkddExp}

\usepackage{amsmath}
\usepackage{pifont}% http://ctan.org/pkg/pifont
\newcommand{\cmark}{\ding{51}}%

\usepackage[numbers, sort, comma, square]{natbib}

\usepackage{adjustbox}
\usepackage{xcolor}
\usepackage{etoolbox}

\usepackage{fourier} 
\usepackage{array}
\usepackage{makecell}
\usepackage{float}
\usepackage[hidelinks]{hyperref}

\newcolumntype{P}[1]{>{\centering\arraybackslash}p{#1}}
\usepackage{tikz}
\usetikzlibrary{shapes,arrows}

\usepackage{mdframed}
\usepackage{lipsum}
\usepackage{multirow}

\usepackage{array}
\newcolumntype{H}{>{\setbox0=\hbox\bgroup}c<{\egroup}@{}}

\newmdtheoremenv{theo}{Theorem}
\usepackage{tabularx}

\usepackage[most]{tcolorbox}

\newtcolorbox[auto counter,number within=subsection]{myBox}[3][]{
arc=2.5mm,
lower separated=false,
fonttitle=\bfseries,
colbacktitle=white!10,
coltitle=black!20!black,
enhanced,
attach boxed title to top left={xshift=0.3cm,
        yshift=-2mm},
colframe=black!20!black,
colback=white%!10,
}

\begin{document}

%%
%% end of the preamble, start of the body of the document source.

\title{Attribution and Obfuscation of Neural Text Authorship:\\
A Data Mining Perspective}

\numberofauthors{3}
%
% You can go ahead and credit authors number 4+ here;
% their names will appear in a section called
% "Additional Authors" just before the Appendices
% (if there are any) or Bibliography (if there
% aren't)

% Put no more than the first THREE authors in the \author command
%%You are free to format the authors in alternate ways if you have more 
%%than three authors.

\author{
%% e-mail address with \email.
\alignauthor Adaku Uchendu \\
       \affaddr{Penn State University}\\
       \affaddr{PA, USA}\\
%       \affaddr{USA}\\
       \email{azu5030@psu.edu}
\alignauthor Thai Le\\
       \affaddr{University of Mississippi}\\
       \affaddr{MS, USA}\\
       %\affaddr{USA}\\
       \email{thaile@olemiss.edu}
\alignauthor Dongwon Lee \\
       \affaddr{Penn State University}\\
       \affaddr{PA, USA}\\
       %\affaddr{USA}\\
       \email{dongwon@psu.edu}
}
\additionalauthors{Additional authors: John Smith (The Th{\o}rvald Group,
email: {\texttt{jsmith@affiliation.org}}) and Julius P.~Kumquat
(The Kumquat Consortium, email: {\texttt{jpkumquat@consortium.net}}).}
\date{30 July 1999}

% \numberofauthors{3}

% \author{

% \alignauthor Adaku Uchendu \\
%        \affaddr{The Pennsylvania State University}\\
%        \affaddr{University Park, PA}\\
%        \affaddr{USA}\\
%        \email{azu5030@psu.edu}
% \alignauthor Thai Le\\
%        \affaddr{The University of Mississippi}\\
%        \affaddr{Oxford, MS}\\
%        \affaddr{USA}\\
%        \email{thaile@olemiss.edu}
% \alignauthor Dongwon Lee \\
%        \affaddr{The Pennsylvania State University}\\
%        \affaddr{University Park, PA}\\
%        \affaddr{USA}\\
%        \email{dongwon@psu.edu}
% }

% \author{Adaku Uchendu}
% \affiliation{%
%   \institution{The Pennsylvania State University}
% %   \streetaddress{University Park}
%   \city{University Park, PA}
% \email{azu5030@psu.edu}

% \author{Thai Le}
% \affiliation{%
%   \institution{The University of Mississippi}
% %   \streetaddress{University Park}
%   \city{Oxford, MS}
% \email{thaile@olemiss.edu}

% \author{Dongwon Lee}
% \affiliation{%
%   \institution{The Pennsylvania State University}
% %   \streetaddress{University Park}
%   \city{University Park, PA}
% \email{dongwon@psu.edu}



\maketitle

\begin{abstract}
% In recent years, Neural Text-Generation (NTG) techniques 
% have greatly advanced, especially
% in the realm of open-ended text generation.
% These NTGs are now able to generate texts that are easily misconstrued as human-written.
% This, therefore, poses a security risk, as such NTGs can be used to generate 
% mis/disinformation at scale with little to no cost. 
% Thus, to combat this novel challenge, 
% several researchers have developed techniques to automatically detect texts generated by NTGs. 
% We will call these texts - \textit{neural texts}. 
% While this niche field of NTD is growing, the field of NTG 
% is growing at a much faster rate, making many of the neural text detectors obsolete. 
% Scaling up the problem further to the case of $k$ NTG methods ($k \geq 2$), each generating uniquely-different yet human-quality texts, two new computational problems emerge: (1) ``Neural''
% \textit{Authorship Attribution} (AA) and (2) ``Neural'' \textit{Authorship Obfuscation} (AO) problems, where
% the AA problem is concerned with attributing the authorship of a given text to one of the $k$ NTG methods, while the AO problem is to evade the authorship of a given text by modifying parts of the text.
% Both problems lie in the intersections between Data Mining, and Security, and their importance and implications are growing rapidly.
% In this survey, therefore, we call-attention to the serious 
% security risk both emerging problems pose
%  and give a comprehensive review of recent literature on the detection and obfuscation of neural text authorships from a Data Mining perspective.
 Two interlocking research questions of growing interest and importance in privacy research are {\it Authorship Attribution} (AA) and {\it Authorship Obfuscation} (AO). Given an artifact, especially a text $t$ in question, an AA solution aims to accurately attribute $t$ to its true author out of many candidate authors while an AO solution aims to modify $t$ to hide its true authorship. Traditionally, the notion of authorship and its accompanying privacy concern is only toward {\em human} authors.
However, in recent years, due to the explosive advancements in Neural Text Generation (NTG) techniques in NLP,  capable of synthesizing human-quality open-ended texts (so-called ``neural texts''), one has to now consider authorships by humans, machines, or  their combination.
Due to the implications and potential threats of neural texts when used maliciously, it has become critical to understand the limitations of traditional AA/AO solutions  and develop novel AA/AO solutions in dealing with neural texts.
In this survey, therefore, we make a comprehensive review of recent literature on the attribution and obfuscation of neural text authorship from a Data Mining perspective, and share our view on their limitations and promising research directions.
\end{abstract}

\section{Introduction} \label{intro}
%% intro AA/AO first -> NTG -> then "neural" AA/AO

%\lee{what is "data mining perspective"? how did Kai's survey paper establish their paper as "data mining perspective"?}

% ---
%\lee{add 1-2 open questions or a paragraph of discussion from DM perspective. It’s our “view” to the DM community.}
\textbf{Natural Language Generation} ({\bf NLG}) is a broad term for AI techniques to produce high-quality human-understandable texts in some human languages, and often encompasses terms such as 
machine translation, dialogue generation, 
text summarization, data-to-text generation,
Question-Answer generation, and 
open-ended or story generation \cite{li2021pretrained, zhang2022survey}. 
Among these, in particular, this survey focuses on the open-ended text generation aspect of NLG.
%because of its ability to generate long coherent articles.
Since the advent of the Transformers architecture in 2018, 
the field of NLG
has experienced exponential improvement. Before 2018, 
leading NLG models were only able to generate 
a few sentences coherently. 
However, after adopting the Transformer architecture into
deep learning-based Language models (LMs), 
NLG models could generate more than a few sentences (i.e., $\geq$ 200 words) coherently.
% GPT-1 was released in 2018 by OpenAI. GPT-1 was 
GPT-1 \cite{radford2018improving} by OpenAI is one of the first such NLG models. 
Since then, many other Transformer-based 
LMs with the capacity to generate long coherent texts have been released (e.g., FAIR~\cite{ng2019facebook,chen2020facebook}, CTRL~\cite{keskar2019ctrl}, PPLM~\cite{dathathri2019plug}, T5~\cite{raffel2020exploring}, WuDao
\footnote{https://github.com/BAAI-WuDao}). 
In fact, as of February 2023, huggingface's \cite{wolf2019huggingface} model repo houses about 8,300 variants of text-generative LMs\footnote{\url{https://huggingface.co/models?pipeline_tag=text-generation}}.
% See Figure \ref{fig:evolution} for growth of such generative LMs.
In this survey, we refer to
these LMs as \textbf{Neural Text Generator} ({\bf NTG})
since they are neural network-based LMs with text-generative abilities. Further, we refer to the texts generated by NTG as {\bf ``neural'' texts}
\footnote{Other names for neural text include
\textit{AI(-generated) text} \cite{kreps2022all},
\textit{Machine(-generated/written) text} \cite{uchendu2020authorship, uchendu2021turingbench,zellers2019defending,gehrmann2019gltr,bakhtin2019real,schuster2020limitations,frohling2021feature, pillutla2021information,galle2021unsupervised,abbasi2008writeprints,tan2020detecting},
% \textit{Machine-made text} \cite{shao2019reverse}, 
\textit{Artificial text} \cite{kushnareva2021artificial}, 
\textit{Computer-generated text} \cite{stiff2022detecting}, 
\textit{Deepfake text} \cite{pudeepfake},
\textit{Auto-generated text} \cite{automik22}, 
and \textit{Synthetic text} 
\cite{diwan2021fingerprinting,munir2021through,brown2020language,guerrero2022synthetic}.}, as opposed to normal texts written by humans as {\bf human texts}.

\begin{figure}
    \centering
    \includegraphics[width=0.8\linewidth]{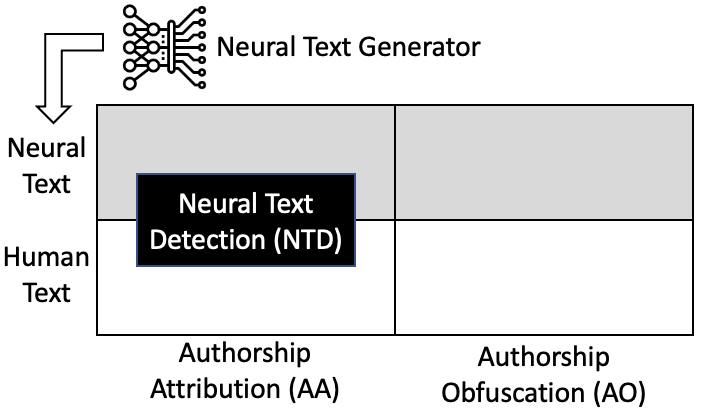}
    \caption{
    The figure illustrates the quadrant of research problems where 
    (1) the \textbf{\underline{\textcolor{gray}{GRAY}}} quadrants are the focus of this survey, and (2) The \textbf{\underline{\textcolor{black}{BLACK}}} box indicates the specialized binary AA problem to distinguish neural texts from human texts. 
    }
    \label{fig:scope}
\end{figure}

\begin{figure}
    \centering
    \includegraphics[width=0.45 \textwidth]{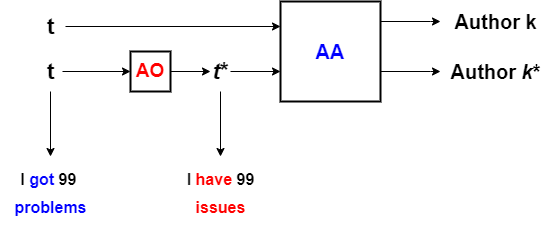}
    \caption{Illustration of both AA and AO problems on neural texts 
    % \thai{can we change X to t and x' to $t^*$, A to k and B to $k^*$ to be consistent with the later definitions?}
    }
    \label{fig:aaao}
\end{figure}

As the qualities of NTGs improve, neural texts become more easily misconstrued as human-written 
\cite{uchendu2021turingbench,zellers2019defending,ippolito2020automatic,gehrmann2019gltr,clark2021all}, 
exacerbating the difficulty 
of distinguishing neural texts from human texts. 
For instance, therefore, such a text generation capability can be misused to  
generate 
misinformation \cite{zellers2019defending,brown2020language,buchanan2021truth}, fake reviews \cite{adelani2020generating} and political
propaganda \cite{varol2017online} 
at scale with little cost. These problems lead to the need to effectively distinguish neural texts from human texts, the so-called {\bf  Neural Text Detection} ({\bf NTD}) problem, which is a sub-problem of a widely studied problem in the privacy community--i.e., authorship attribution.
In fact, two interlocking research questions in privacy research, heavily studied but of growing interest,  are {\bf Authorship Attribution} ({\bf AA}) and {\bf Authorship Obfuscation} ({\bf AO}). Given an artifact, especially a text $t$ in question, an AA solution aims to accurately attribute $t$ to its true author out of $k$ candidate authors while an AO solution aims to modify $t$ to hide its true authorship.
Therefore, NTD is a specialized case, a Turing Test, of AA with $k=2$ authors (i.e., human vs. machine). Figure~\ref{fig:scope} illustrates the quadrant of research problems, while Figure~\ref{fig:aaao} illustrates  how both AA and AO  problems work hand-in-hand.
%Our survey bears similarity with another survey for NTD \cite{jawahar2020automatic} published in 2020.\lee{but how are two different? if only similar, why another survey?}
%In recent years, both NTG and NTD have greatly improved. Firstly, the AO problem for neural texts was only briefly introduced in \cite{wolff2020attacking}, and since then, it has been more thoroughly studied. Thus, we are the first NTD survey that comprehensively discusses NTD through the AA and AO landscape. 

Traditionally, the notion of authorship and its accompanying privacy concern in both AA and AO problems are only toward {\em human} authors.
However, with the arrival of generative AI technologies and due to the potential threats of misused neural texts,
one now has to consider authorships by humans, machines, or  their combination, and 
re-thinks about effective solutions for both AA and AO problems for neural texts.
Hence, to guide these 
developments, in this survey, we provide a detailed analysis of both AA and AO 
problems, their existing solutions, and our perspective on the open challenges.
As both AA and AO problems are essentially computational learning problems, 
we discuss the landscape from 
\textit{A Data Mining Perspective} and call attention to the security challenges that need to be solved.
We believe that the issues of these novel AA/AO problems for neural texts 
are ``nuanced'' and therefore require nuanced solutions 
from the Data Mining and Machine Learning community. 
%See Figure \ref{fig:scope} for the scope of this survey. 

\begin{comment}
Lastly, this paper will be arranged in the following way: 
Section \ref{definition} will briefly discuss the 
state-of-the-art NTGs, and data labeling process;
% Section \ref{trad-AA} will briefly discuss the traditional AA problem and the difference between it and AA tasks for neural text detection;
Section \ref{AA} will discuss the AA models for NTD 
as well as the taxonomy for the AA models for neural texts;
Section \ref{authorshipobf} will discuss in detail the  
AO techniques used to obfuscate neural texts and taxonomy for 
AO techniques used on neural texts;
Sections \ref{open} and \ref{app} discuss the current open problems in 
NTD, and applications to real life, respectively. 
Finally, Section \ref{conclusion} will briefly outline 
the landscape of the AA \& AO problems discussed in the survey. 
\end{comment}

\section{Neural Text Generation} \label{definition}

\begin{figure}
    \centering
    \includegraphics[width=.49\textwidth]{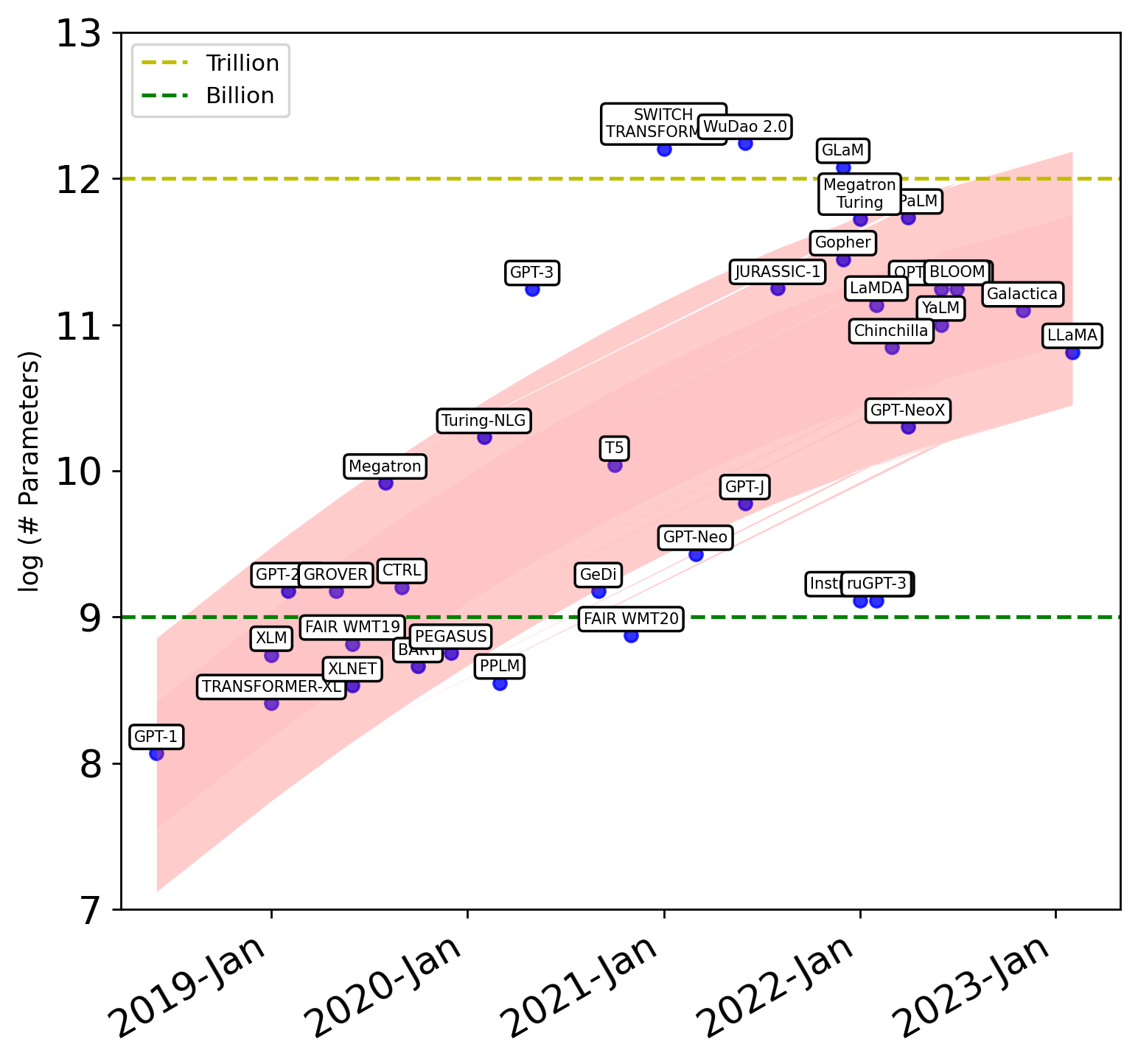}
    \caption{
    % \lee{redraw to look more professional/formal. Y-axis should be "The size of model parameter (in log)". on X-axis, let's only use year. use black color, not gray, on X and Y axis as currently not clear. remove title above}
    Evolution of Neural Text Generators (NTGs) from 2018 to 2023 ($Y$-axis is a log plot of \# of parameters). 
    % We extended the plot created by \cite{uchendu2021turingbench} to include NTGs released in 2022.
    }
    \label{fig:evolution}
    \vspace{-10pt}
\end{figure}

\begin{table*}[t!]
\centering
\resizebox{16cm}{!}{\begin{tabular}{|c|c|c|}
\hline

% \textbf{NTG} & \textbf{Author}  & \begin{tabular}[c]{@{}c@{}} \textbf{Description} \end{tabular}   \\
\multicolumn{1}{|c|}{\textbf{NTG}} & \multicolumn{1}{c|}{\textbf{Author}} & \multicolumn{1}{c|}{\textbf{Description}} \\

\hline

GPT-1 \cite{radford2018improving}   & OpenAI &
\begin{tabular}[c]{@{}c@{}}
It used Transformers to model a simple concept - to predict the next token, given 
the previous token. 
\end{tabular}
\\ \hline

GPT-2  \cite{radford2019language} &  OpenAI & 
\begin{tabular}[c]{@{}c@{}}
GPT-1 scaled up. There are 4 GPT-2 pre-trained models -  
\textit{small} (124 million parameters), \\ \textit{medium} (355 million parameters),
\textit{large} (774  million parameters), and \textit{x-large} (1558 million  parameters)   
\end{tabular}
\\ \hline

GPT-3 \cite{brown2020language} &  OpenAI & 
\begin{tabular}[c]{@{}c@{}}
GPT-2, scaled up - increasing parameter and train data size. 
\end{tabular}
\\ \hline

GROVER  \cite{zellers2019defending}  & AllenAI & 
\begin{tabular}[c]{@{}c@{}}
Similar to GPT-2 architecture and trained to generate political news. 
There are 3 \\pre-trained models: \textit{GROVER-base}, \textit{GROVER-large}, \textit{GROVER-mega} 
\end{tabular}
\\ \hline

CTRL \cite{keskar2019ctrl} & Salesforce & 
\begin{tabular}[c]{@{}c@{}}
Conditional Transformer LM For controllable generation uses control codes to guide generation 
\end{tabular}
\\ \hline

XLM \cite{lample2019cross} & Facebook & 
\begin{tabular}[c]{@{}c@{}}
A Cross-lingual Language Model trained on various languages. Only the English model is used for AA  
\end{tabular}
\\ \hline

XLNET \cite{yang2019xlnet} & Google &  
\begin{tabular}[c]{@{}c@{}}
A generalized auto-regressive pre-training method that adopts the Transformer-XL framework 
\end{tabular}
\\ \hline

FAIR\_wmt \cite{ng2019facebook,chen2020facebook} & Facebook &
\begin{tabular}[c]{@{}c@{}}
FAIR\_wmt has 3 language models - English, Russian, and German. Only the English \\
model\footnote{http://shorturl.at/swDHJ} is used, which has 2 models - \textit{WMT19} \cite{ng2019facebook}  and  \textit{WMT20} \cite{chen2020facebook}.  
\end{tabular}
\\ \hline

TRANSFORMER\_XL \cite{dai2019transformer} & Google & 
\begin{tabular}[c]{@{}c@{}}
Another Transformer model that learns long-term dependency to improve long coherent text generation
\end{tabular}
\\ \hline

PPLM \cite{dathathri2019plug} & Uber & 
\begin{tabular}[c]{@{}c@{}}
The Plug and Play Language Models (PPLM) model upon
GPT-2 by fusing the GPT-2 medium with a bag of \\words (BoW) models. These BoW models are \textit{legal, military, monsters, politics, positive\_words, religion,} \\\textit{science, space, technology}.
PPLM can plug in any GPT-2 pre-trained model to generate texts 
\end{tabular}
\\ \hline

Switch Transformer \cite{fedus2022switch} & Google & 
\begin{tabular}[c]{@{}c@{}}
Google uses a 
switch Transformer to build a sparse neural LM with 1.6T 
parameters are built 
\end{tabular}
\\ \hline

GPT-Neo \cite{gao2020pile} & EleutherAI & 
\begin{tabular}[c]{@{}c@{}}
EleutherAI replicates GPT-3's architecture.
There a 2 model sizes - 1.3B and 2.7B parameters  
\end{tabular}
\\ \hline

GPT-NeoX \cite{black2022gpt} & EleutherAI &  
\begin{tabular}[c]{@{}c@{}}
A 20 billion parameter autoregressive replication of GPT-3. 
\end{tabular}
\\ \hline

GPT-J \cite{gpt-j}  & EleutherAI & 
\begin{tabular}[c]{@{}c@{}}
A 6B parameter model similar to the GPT-Neo and GPT-NeoX that uses Mesh Transformer JAX \cite{meshtransformerjax} \\ framework to train the model with Pile\footnote{https://pile.eleuther.ai/} dataset, a large curated dataset created by EleutherAI 
\end{tabular}
\\ \hline

T5 \cite{raffel2020exploring} & Google & 
\begin{tabular}[c]{@{}c@{}}
An encoder-decoder text-to-text Transformer-based model. \\T5 has 5 pre-trained models - \textit{T5-small, T5-base, T5-large, T5-3b,} and \textit{T5-11b}  
\end{tabular}
\\ \hline

BART \cite{lewis2020bart} & Facebook & 
\begin{tabular}[c]{@{}c@{}}
This is another encoder-decoder Transformer-based LM, most effective when fine-tuned
\end{tabular}
 \\ \hline

PaLM \cite{chowdhery2022palm} & Google & 
\begin{tabular}[c]{@{}c@{}}
PaLM stands for Pathways Language Model. It is a dense decoder-only Transformer-based \\ model trained with \cite{barham2022pathways}'s pathways system framework  
\end{tabular}
\\ \hline

OPT-175B \cite{zhang2022opt} & Meta & 
\begin{tabular}[c]{@{}c@{}}
Meta's response to GPT-3. OPT-175B uses a similar framework to GPT-3 \\ but the training costs 1/7th the carbon footprint of GPT-3. 
\end{tabular}
\\ \hline

GeDi \cite{krause2021gedi} & Salesforce & 
\begin{tabular}[c]{@{}c@{}}
GeDi stands for Generative Discriminator Guided
Sequence Generation. Similar to PPLM, \\GeDi controls text generation using small LMs as generative discriminators  
\end{tabular}
\\
\hline
\end{tabular}}
\caption{Description of state-of-the-art Neural Text Generators (NTGs).}
\label{tab:desc}
\end{table*}

We first select a handful of Neural Text Generator (NTG) that we focus on in this survey, and introduce a list of popular datasets with neural texts. 

% \begin{figure}[ht]
%     \centering
%     \includegraphics[width=1\linewidth]{figures/NTGprocess.drawio.png}
%     \caption{Illustration of how NTGs generate texts}
%     \label{fig:ntgprocess}
% \end{figure}

\subsection{Neural Text Generators (NTGs)}
Those NTGs studied in this survey are large-scale probabilistic LMs that are capable of generating long-coherent texts (e.g., $\geq$ 200 words).
These LMs are trained on massive amounts of unstructured texts.
Based on its architecture structure (e.g., encoder-decoder or 
decoder only), these LMs use a prompt, a snippet of human-written text, to guide the generation of
texts, emulating the most similar style from the training set and predicting one token at a time.
%See Figure \ref{fig:ntgprocess} for an illustration of how neural text data is generated. 
% \lee{Can we describe on the basics of text generation by LMs here?} 
%There are different kinds of LMs that can generate texts (GAN, LSTM, etc.), however, only Transformer-based models 
%that can generate long-coherent texts ($\geq$ 200 words).
% Within Transformer-based NTGs, there are different kinds of models - 
% Encoder-Decoder, and Decoder-only. 
%Discussing NTGs in detail can easily become its own survey paper,  therefore, we only give a very brief description of the NTGs that will be mentioned. 
Recent works such as \cite{zhang2022survey,li2021pretrained} survey these NTGs in detail.
% In order not to veer far from the scope of this survey,  
% we only give a very brief description of the NTGs that will be mentioned. 
The progress of NTGs in recent years has been expeditious.
%exponential with respect to their parameter sizes. %currently with more than 8,200 text-generative LMs, housed in huggingface's model hub. 
As shown in Figure \ref{fig:evolution}, for instance, 
the sizes of NTGs with respect to their parameters are growing at an exponential rate,
yielding the rapid improvement in the quality of neural texts, thus exacerbating the AA/AO problems. 
Table \ref{tab:desc} shows a summary of state-of-the-art NTGs, where many entries are drawn from \cite{uchendu2020authorship,uchendu2021turingbench}.
%Note that majority of the entries in the table was gotten from related papers such as \cite{uchendu2020authorship,uchendu2021turingbench}.

Within LMs, in particular, hyperparameters matter a great deal when generating texts. 
The choice of these hyperparameters,  referred to as \textit{decoding strategies}, greatly affects the quality of generated neural texts. According to \cite{holtzman2019curious}, 
there are 6 \textit{decoding strategies}:
(1) \textit{Greedy sampling} selects the best probable word,
(2) \textit{Random sampling} does a stochastic search for a sufficient word, 
(3) \textit{Top-k sampling} samples from top-k most probable words,
(4) \textit{Beam search} searches for most probable candidate sequences,
(5) \textit{Nucleus (Top-p) sampling} samples similar to top-k, 
but its focus is on the smallest possible set of top words, such that the sum of their probabilities is $\geq$ p, and
(6) \textit{Temperature} scales logits to either increase or decrease the entropy of sampling.
%(0 temperature=max likelihood, infinite temperature=uniform sampling).
NTG models often use 
\textit{Top-k sampling}, \textit{Beam search}, 
\textit{Nucleus (Top-p) sampling}, and \textit{Temperature} decoding strategies as they produce higher quality texts than other decoding 
strategies. In fact, 
\cite{holtzman2019curious} reports
that neural texts generated with the
\textit{Nucleus (Top-p) sampling} strategy are more challenging to attribute authorships.

\subsection{Neural Text Datasets} \label{data}
% \subsection{Data Description}

To investigate both AA/AO problems for neural texts, one needs benchmark datasets of neural texts.
%One of the biggest difficulties in NTD is the lack of sufficient training data.
%This is mostly due to the computational cost of generating a 
%large number of neural texts 
%needed to train AA models well. 
% We will discuss all the challenges of generating data in 
% the next subsection. 
%However, some decently sized data for the AA task have been generated. See 
Table \ref{tab:data}  describes a list of publicly available datasets that contain neural texts. 
Labels of the texts in the datasets are either binary (neural vs. human text) or multi-class (having multiple neural texts generated by different NTGs vs. 1 human label). The majority of recent studies have focused on the binary case of the Turing Test to check if a given text is written by a human author or machine (i.e., one of NTGs).
%Researchers have mostly studied the AA special case -\textbf{Turing Test} (where $k=2$), so there is more application of the binary datasets in the literature. Only publicly available datasets were included in the table. 
%Due to the nature of the task, authorship attribution of neural texts can be posed as either a multi-class or a binary classification problem. Hence, 
Researchers utilized clever labeling 
and generation methods to build datasets:
(1) \textit{Binary dataset}:  Researchers first collect human-written texts (e.g., news, 
blogs, stories, or recipes) and use snippets of these
texts as prompts to the chosen NTG to generate a machine-written (neural)  text.
%similar version of the document with the NTGs. This creates a 2-labeled dataset -  human vs. machine.
(2) \textit{Multi-class dataset}: Starting with human-written texts as prompts, this generates multiple neural texts by using different
NTG architectures 
(i.e., human vs. GPT-1, GROVER, PPLM, etc.), 
 different pre-trained sizes of the same NTG 
architecture (i.e., human vs. GPT-2 small vs. GPT-2 medium vs. GPT-2 
large vs. GPT-2 XL, etc.),
 different decoding strategies (i.e., human vs. GPT-2  top-k vs. GPT-2 top-p, etc.).

\begin{table*}[t!]
\centering
\resizebox{14cm}{!}{
\begin{tabular}{|c|c|c|c|cH|}
\hline

\textbf{Name}  & 
        \begin{tabular}[c]{@{}c@{}}
\textbf{Description} \end{tabular} & 
     \textbf{Category} & \textbf{Domain} &
       \textbf{Labels} & \textbf{Adopter} \\
% \toprule
        \hline

        \begin{tabular}[c]{@{}c@{}}   
        GPT-2 dataset \\ \cite{radford2018improving} 
        \end{tabular}
        &  
        \begin{tabular}[c]{@{}c@{}}
        250K Webtext (Human dataset) 
         vs. 250K GPT-2 \\ (small, medium, large, \& XL) 
        %  \footnote{https://github.com/openai/gpt-2-output-dataset} 
        \end{tabular}
        & Binary & News & GPT-2 \& Human & \cite{wolff2020attacking,zhong2020neural,kushnareva2021artificial,pillutla2021information,crothers2022adversarial,stiff2022detecting,pudeepfake} \\

        \hline
        \begin{tabular}[c]{@{}c@{}}
        GROVER \\dataset \cite{zellers2019defending} 
        \end{tabular} & 
        \begin{tabular}[c]{@{}c@{}}
        Using April 2019 news articles as the prompt, \\GROVER-Mega generated news articles 
        \end{tabular}
        % \footnote{https://github.com/rowanz/grover/tree/master/generation\_examples} 
        &
        Binary & News &  GROVER \& Human & 
        \cite{zhong2020neural,kushnareva2021artificial,bhat2020effectively,zellers2019defending,pillutla2021information,stiff2022detecting,puunraveling,pudeepfake}  \\
        
        \hline 
        
        \begin{tabular}[c]{@{}c@{}}
        TuringBench-\\AA \cite{uchendu2021turingbench} 
        \end{tabular}
        & 
        \begin{tabular}[c]{@{}c@{}}
        Used 10K human-written news articles (mostly \\Politics) from
        CNN, etc. to generate 10K articles\\
        each from 19 NTGs.   
        \end{tabular}
        % \footnote{https://huggingface.co/datasets/turingbench/TuringBench} 
        & 
        Multi-class & News &
        \begin{tabular}[c]{@{}c@{}}
         Human \& 19 Machine\\ labels 
        (GPT-1, GPT-2 \\small, etc.)
        \end{tabular}
        % GPT-2 medium, GPT-2 large, GPT-2 XL, GPT-2 PyTorch, GPT-3, GROVER base, GROVER large, GROVER mega, CTRL, XLM, XLNET base, 
        % XLNET large, FAIR\_wmt19, FAIR\_wmt20, Transformer\_XL, 
        % PPLM\_distil, PPLM\_gpt2) 
        & \cite{uchendu2021turingbench,constrabert,alison}  \\
        
        \hline
        
        \begin{tabular}[c]{@{}c@{}}
        TuringBench-\\TT \cite{uchendu2021turingbench} 
        \end{tabular}
        & 
        \begin{tabular}[c]{@{}c@{}}
        The same dataset as \textit{TuringBench-AA}, except that \\the datasets are 19 versions of human vs. each of \\ the 19 NTGs.  
        \end{tabular}
        % \footnote{https://huggingface.co/datasets/turingbench/TuringBench} 
        & 
        Binary & News &
        \begin{tabular}[c]{@{}c@{}}
         19 Human vs. Machine \\combinations (GPT-2, etc.) 
        \end{tabular}
        &  \cite{uchendu2021turingbench}  \\
        
        \hline
        \begin{tabular}[c]{@{}c@{}}
        Authorship \\Attribution-\\AA \cite{uchendu2020authorship} 
        \end{tabular}
        & 
        \begin{tabular}[c]{@{}c@{}}
        Used 1K human-written articles to generate 1K \\articles each from  8 Artificial Text Generators 
        \end{tabular}
    % \footnote{https://huggingface.co/datasets/turingbench/TuringBench} 
      & 
       Multi-class & News &
      \begin{tabular}[c]{@{}c@{}}
         1 human vs 8 Machine\\ labels (GPT-1, GPT-2, etc.) 
    \end{tabular} &
        \cite{uchendu2020authorship,stiff2022detecting} \\
        
        \hline
        \begin{tabular}[c]{@{}c@{}}
        Authorship \\Attribution-\\TT \cite{uchendu2020authorship} 
        \end{tabular}
        &
        \begin{tabular}[c]{@{}c@{}}
        The same dataset as 
        \textit{Authorship Attribution-AA} \\except that the 
        datasets are 8 versions of human vs. \\each of the 8 NTGs
        \end{tabular}
        % \footnote{https://huggingface.co/datasets/turingbench/TuringBench} 
        & 
        Binary & News &  
        \begin{tabular}[c]{@{}c@{}}
        8 humans vs Neural \\combinations  
        \end{tabular}
        & 
        \cite{uchendu2020authorship,stiff2022detecting}  \\
        
        \hline 
        
        \begin{tabular}[c]{@{}c@{}}
        Authorship \\Attribution-\\TT mix \cite{uchendu2020authorship}
        \end{tabular}
        & 
        \begin{tabular}[c]{@{}c@{}}
        The same dataset as 
        \textit{Authorship Attribution-AA}\\ except that the 
        dataset is human vs. Machine\\ (which is a mixture of all the 8 NTGs) 
        \end{tabular}
        % \footnote{https://huggingface.co/datasets/turingbench/TuringBench} 
        & 
        Binary & News &  
        \begin{tabular}[c]{@{}c@{}}
        1 human vs Machine\\ (8 different machines) 
        \end{tabular}
        & 
        \cite{uchendu2020authorship} \\
        
        \hline

        \begin{tabular}[c]{@{}c@{}}
        Academic \\Papers \\ \& Abstracts 
        \cite{liyanage2022benchmark} 
        \end{tabular}
        & 
        \begin{tabular}[c]{@{}c@{}}
        2 datasets - 
        (1) FULL: using a short prompt for a \\
        human-written paper generated an academic \\
        paper using GPT-2; 
        (2) PARTIAL: Replacing \\
        sentences from an Abstract with Arxiv-NLP model \\
        generations 
        \end{tabular}
    %  \footnote{https://github.com/vijini/GeneratedTextDetection}
     & 
        Binary & 
        \begin{tabular}[c]{@{}c@{}}
        Academic \\Abstracts 
        \end{tabular}
        &
        \begin{tabular}[c]{@{}c@{}}
         Human \& Machine 
        \end{tabular}
        & 
        \cite{liyanage2022benchmark} \\
        
        \hline

        \begin{tabular}[c]{@{}c@{}}
        Hybrid Human-\\Machine Text \\ \cite{cutlerautomatic} 
        \end{tabular}
        & 
        \begin{tabular}[c]{@{}c@{}}
        Using human-written text in domains - News, \\Reddit, and Recipes to generate continuations of\\ the text using GPT-2 XL
        \end{tabular}
        % \footnote{https://drive.google.com/file/d/1YlG1O\_w\_fVTc9oLkoz9eBlmzHFuGfNze/view} 
         & Binary & 
        \begin{tabular}[c]{@{}c@{}}
          News,\\ Reddit,\\ Recipes 
          \end{tabular}
          
          &
        \begin{tabular}[c]{@{}c@{}}
         Human prompt \& \\Machine texts 
         \end{tabular}
         & \cite{cutlerautomatic}  \\
        
        \hline

        \begin{tabular}[c]{@{}c@{}}
        Amazon \\Reviews \cite{adelani2020generating} 
        \end{tabular}
        & 
        \begin{tabular}[c]{@{}c@{}}
        Fine-tuned GPT-2 on 3.6 M Amazon and 560K\\
        Yelp reviews  
        \end{tabular}
        & Binary & Reviews & 
        Human \& Machine  
         & \cite{adelani2020generating,kushnareva2021artificial} \\
         
         \hline

        \begin{tabular}[c]{@{}c@{}}
         Human-Machine 
         \\Pairs \cite{puunraveling} 
        \end{tabular}
        %  \footnote{https://drive.google.com/file/d/1xA9TtDYJE9BE
        %  wecL8QJ5d0LTytn5hhBr/edit} 
         & 
         \begin{tabular}[c]{@{}c@{}}
         Generated texts with GROVER mega and GPT-2 \\XL
         with top-p decoding strategy. Paired \\ human-written texts with a similar neurally \\generated version 
         \end{tabular}
         & 
        \begin{tabular}[c]{@{}c@{}}
         Binary \\
         (Human-\\Machine\\ pairs) 
        \end{tabular} 
         & 
        \begin{tabular}[c]{@{}c@{}}
         Online \\forums \\ \& News 
         \end{tabular}
         &
         Human \& Machine & \cite{puunraveling}\\
         
         \hline
                 
        \begin{tabular}[c]{@{}c@{}}
         NeuralNews \\dataset \cite{tan2020detecting}
         \end{tabular}
        %  \footnote{https://github.com/rxtan2/DIDAN/tree/main/data} 
         & 
         \begin{tabular}[c]{@{}c@{}}
         Using the GoodNews \cite{biten2019good} dataset as the \\human-written prompt to generate texts with \\ GROVER.
         Real images are included in each of the \\articles.
         \end{tabular}
        %  The Real images were captioned by fake and real captions.
        %  Also, using an entity-aware image captioning model \cite{biten2019good}, captions for these images were generated. 
        %  This resulted in 4 article types: 
        %  \textbf{(A)} \textit{Real Articles and Real Captions}; 
        %  \textbf{(B)} \textit{Real Articles and Generated Captions};
        %  \textbf{(C)} \textit{Generated Articles and Real Captions};
        %  \textbf{(D)} \textit{Generated Articles and Generated Captions}.
         & Binary & News & Human \& Machine  & \cite{tan2020detecting} \\

         \hline 
        \begin{tabular}[c]{@{}c@{}}
         SynSciPass \cite{rosati2022synscipass} 
         \end{tabular} &
          \begin{tabular}[c]{@{}c@{}}
        Built dataset using 3 
        potential sources of neural text: \\(1)
        open-ended text generators
        like GPT-2 \& BLOOM \\ (2) paraphrase models like SCIgen and PEGASUS
        and \\(3) translation models like Spinbot, real, \\Google translate, and Opus.
         \end{tabular}
         & Multi-class & 
         
         \begin{tabular}[c]{@{}c@{}}
         scientific \\articles 
        \end{tabular} &

        \begin{tabular}[c]{@{}c@{}}
         generate, translate, \\paraphrase, \& human
        \end{tabular} 
        & \cite{rosati2022synscipass} \\

        \hline 
        TweepFake \cite{fagni2021tweepfake} &
        \begin{tabular}[c]{@{}c@{}}
        Collected tweets generated by Twitter bots \\
        and grouped them into tweets generated by \\
        GPT-2, RNN, and other bots
         \end{tabular} &
         Binary & Tweets & 
         Human \& Machine &
         \cite{fagni2021tweepfake} \\
         
    %   \bottomrule
    \hline
    \end{tabular}
    }
    % \end{adjustbox}

    \caption{Datasets with neural texts}
    \label{tab:data}
    \vspace{-10pt}
\end{table*}

\renewcommand{\tabcolsep}{2pt}
\begin{table*}[th!]
\centering
\resizebox{14cm}{!}{
\begin{tabular}{|c|c|c|c|c|c|}
% \toprule
\hline
        \textbf{Model Name} & \textbf{Classifier Type} & \textbf{Category} & \textbf{Learning Type} & \textbf{Interpretable} & \textbf{Training dataset}\\
       
        \hline
       GROVER detector \cite{zellers2019defending} & DL (Transformer-based) & Binary & Supervised &  & GROVER \\
       
       \hline
       GLTR \cite{gehrmann2019gltr} & Statistical & Binary & Unsupervised & \cmark & GPT-2 
        \\
       
       \hline
       GPT-2 detector \cite{gpt2outputdetector} & DL (Transformer-based) & Binary & Supervised &  & GPT-2 \\
       
       \hline 
       OpenAI detector \cite{gpt2outputdetector} & DL (Transformer-based) & Multi-class & Supervised &  & 
       \begin{tabular}[c]{@{}c@{}}
       GPT-2 \& TuringBench-AA  
       \end{tabular}
       \\
       
       \hline
       RoBERTa-TT \cite{uchendu2021turingbench} & DL (Transformer-based) & Binary & Supervised &  & TuringBench-TT \\
       
       \hline
       BERT-TT \cite{uchendu2021turingbench} & DL (Transformer-based) & Binary & Supervised &  & TuringBench-TT  \\
       
       \hline
       RoBERTa-Multi \cite{uchendu2021turingbench}  & DL (Transformer-based) & Multi-class & Supervised &  & TuringBench-AA \\
       
       \hline
       BERT-Multi \cite{uchendu2021turingbench}  & DL (Transformer-based) & Multi-class & Supervised &  & TuringBench-AA \\
       
       \hline
       TDA-based detector \cite{kushnareva2021artificial} & Hybrid & Binary & Supervised & \cmark & GPT-2 \\
       
       \hline
       FAST  \cite{zhong2020neural} & Hybrid & Binary & Supervised &  & GPT-2 \& GROVER  \\
       
       \hline
       Energy discriminator \cite{bakhtin2019real} & DL (Energy-based) & Binary & Supervised &  & GROVER \\
       
       \hline 
       MAUVE \cite{pillutla2021information} & Statistical & Binary & Unsupervised & \cmark & GPT-2 \& GROVER \\
       
       \hline
       Distribution detector \cite{galle2021unsupervised} & Statistical & Binary & Unsupervised & \cmark & GPT-2 \\
       
       \hline
       \begin{tabular}[c]{@{}c@{}}
       Feature-based detector \cite{frohling2021feature} 
       \end{tabular}
       & Stylometric & Binary & Supervised & \cmark & 
       \begin{tabular}[c]{@{}c@{}}
       GPT-2, GPT-3, \& GROVER 
       \end{tabular}
       \\
       
       \hline
       Linguistic model \cite{uchendu2020authorship} & Stylometric & Multi-class & Supervised & \cmark & 
       \begin{tabular}[c]{@{}c@{}}
       Authorship Attribution-AA 
       \end{tabular}
       \\
       
       \hline
       DIDAN \cite{tan2020detecting} & Hybrid & Binary & Supervised &  & 
       \begin{tabular}[c]{@{}c@{}}
       Fake images w/ Human \\ vs. GROVER news 
       \end{tabular}
       \\
       
       \hline 
       XLNet-FT \cite{munir2021through} & DL (Transformer-based) & Multi-class & Supervised &  & 
       \begin{tabular}[c]{@{}c@{}}       
       GPT-1, GPT-2, XLNet, \\ 
        \& BART Reddit posts 
       \end{tabular}
       \\
       
       \hline 
       Constra-DeBERTa  \cite{constrabert} & DL (Transformer-based) & Multi-class & Supervised &  & TuringBench  \\
       
       \hline 
       Fingerprint detector  \cite{diwan2021fingerprinting} & Hybrid &  Multi-class & Supervised &  & 
       \begin{tabular}[c]{@{}c@{}}
       GPT-2 bot subreddit posts 
       \end{tabular}
       \\
    %   \footnote{https://www.reddit.com/r/SubSimulatorGPT2Meta/} \\
       
       \hline 
       DistilBERT-Academia \cite{liyanage2022benchmark} & DL (Transformer-based) & Binary & Supervised &  & 
       \begin{tabular}[c]{@{}c@{}}
       GPT-2 Academia \\abstract \& paper  
       \end{tabular}
       \\
       
       \hline
       \begin{tabular}[c]{@{}c@{}}
       Sentiment modeling detector \cite{adelani2020generating} 
       \end{tabular}
       & DL (Glove-based) & Binary & Supervised &  & GPT-2 Amazon Reviews  \\
       
       \hline 
       
       BERT-Defense \cite{ippolito2020automatic} & DL (Transformer-based) & Binary & Supervised & & GPT-2 Large WebText  \\
       
       \hline
       RoBERTa-Defense \cite{pudeepfake} & DL (Transformer-based) & Binary & Supervised & & GROVER  \\

       \hline
       RoBERTa w/ GCN \cite{jawahar2022automatic} & DL (Transformer-based) & Binary & Supervised & & GPT-2  \\

       \hline 
       DeBERTa v3 \cite{rosati2022synscipass} & DL (Transformer-based) & Multi-class & 
       Supervised & & 
       \begin{tabular}[c]{@{}c@{}}
       GPT-2, BLOOM, PEGASUS, \\OPUS, SCIgen, Spinbot
       \end{tabular} \\

       \hline 
       Ensemble \cite{gambini2022pushing} & DL (Transformer-based) &
       Binary & Supervised &  &
       TweepFake \\

        \hline 
        CoCo \cite{liu2022coco} & Hybrid & Binary & Supervised & \cmark & 
        \begin{tabular}[c]{@{}c@{}}
       GPT-2 \& GROVER
       \end{tabular} \\

       \hline 
       DetectGPT \cite{mitchell2023detectgpt} & Statistical & Binary &
       Unsupervised & \cmark &
        \begin{tabular}[c]{@{}c@{}}
       GPT-2, OPT-2.7, GPT-Neo-2.7, \\ GPT-J, \& GPT-NeoX
       \end{tabular} \\
       
       \hline
    \end{tabular}
    }
    \caption{Authorship Attribution models (Binary \& Multi-class) for NTD }
    \label{tab:detectors}
    \vspace{-10pt}
\end{table*}
% \begin{figure*}
%     \centering
%     \includegraphics[width=0.8\linewidth]{figures/taxonomy_AA.drawio.png}
%     \caption{Taxonomy of Authorship Attribution models for Neural Text Detection}
%     \label{fig:AATax}
% \end{figure*}

\begin{figure}
    \centering
    \includegraphics[width=0.75 \linewidth]{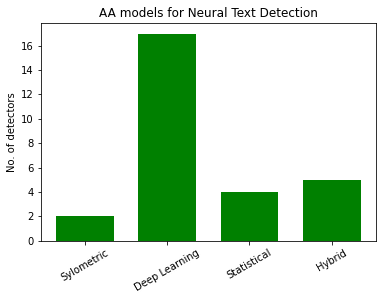}
    \caption{Number of AA solutions for NTD per category in the Taxonomy}
    \label{fig:aa_bar}
\end{figure}

\section{Authorship Attribution for \\Neural Texts} \label{AA}
Traditional AA problem studies the attribution of an author to a piece of written text out of a number of possible authors. However, in the literature, researchers have also studied a few variations of the AA problem. For instance, the  \textit{Author Verification (AV)} problem 
\cite{kestemont2021overview,kestemont2021overview,stamatatos2016authorship,stamatatos2022overview,bevendorff2022overview,tyo2022state}
% \lee{cite a few examples of AV} 
studies if the given two texts, $t_1$ and $t_2$, are written by the same author?
With the rise of neural texts, in addition, a specialized case of AA problem, NTD \cite{zellers2019defending,gehman2020realtoxicityprompts,uchendu2020authorship,bakhtin2019real,gpt2outputdetector,tan2020detecting}, 
% \lee{cite a few examples too} 
studies:
% \begin{myBox}[]{Definition}{AO}
% if the given text, $t$, is written by a human or NTG?
% \end{myBox}
By and large, however, a good solution for the standard AA problem can  lead to a good solution for other variations of the AA problem. As such, we focus on the survey of the standard AA problem for neural texts. Thus, our paper formally defines the AA task as follows.

\begin{myBox}[]{Definition}{AA}
\textsc{\textbf{Definition of AA for Neural Texts.}} \textit{Given a text $t$, the AA model $F(x)$ attributes the text $t$ to its true author $k$--i.e., $k{=}F(x)$, which can be either a human or an NTG author.}
\end{myBox}
% In the literature, researchers have studied several variations of AA problem as follows:
% % There are 4 variations of the AA problem studied:
% \begin{itemize}
%     \item[P1:] Are given two texts, $t_1$ and $t_2$, 
%      written by 
%     the same author? (\textit{Author Verification})
    
%     \item[P2:] Is the given text, $t$, written by human or an NTG? (NTD)
    
%     \item[P3:] Given a text, $t$, attribute $t$ to its true author out of $k$ candidate  (human or machine) authors  (\textit{Standard Authorship Attribution})
    
%     \item[P4:] Given a pair of human \& neural texts, $T1-T2$, which of one of the pair - $T1$ or $T2$ is authored by a neural method?
%     The goal here is to assign a higher machine probability to the neural texts than to the human-written ones.
% \end{itemize}

% P1 is a binary classification problem and has only been studied by one author 
% \cite{uchendu2020authorship}. Based on the results, 
% we conclude that P1 is the most trivial 
% variation of the AA problems. 
% It has been shown to be a somewhat trivial AA task to solve, especially 
% when the dataset is balanced.
% P2 is the most studied variation of 
% the AA problem. 
% P3 is the second most studied variation and the more realistic scenario in real life. 
% P4 has only been studied by 
% \cite{zellers2019defending, zhong2020neural}. Compared to the other variations 
% P4 is the least realistic scenario in real life. 

\tikzstyle{rect} = [draw, rectangle, rounded corners, fill=red!20, 
text width=6em, text centered, minimum width=2em]
\tikzstyle{rect1} = [draw, rectangle, rounded corners, fill=gray!20, 
text width=10em, text centered, minimum width=2em]
\tikzstyle{rect2} = [draw, rectangle, rounded corners, fill=green!20, text width=10em, text centered, minimum width=10em]
\tikzstyle{rect3} = [draw, rectangle, rounded corners, fill=red!20, text width=10em, text centered, minimum width=10em]
\tikzstyle{rect4} = [draw, rectangle, rounded corners, fill=blue!20, text width=10em, text centered, minimum width=10em]
\tikzstyle{rect5} = [draw, rectangle, rounded corners, fill=orange!20, text width=10em, text centered, minimum width=10em]
\tikzstyle{rect6} = [draw, rectangle, rounded corners, fill=yellow!20, text width=10em, text centered, minimum width=10em]

\tikzstyle{elli} = [draw, ellipse, fill=white!20,
minimum height=2em]
\tikzstyle{circ} = [draw, circle, fill=white!20, minimum width=8pt,
inner sep=10pt]
\tikzstyle{diam} = [draw, diamond, fill=white=!20, text width=6em,
text badly centered, inner sep=0pt]
\tikzstyle{line} = [draw, -latex']

\begin{figure}[htb!]
\begin{center}
\begin{tikzpicture}[node distance = 1.8cm, scale=0.75, transform shape]
\node [rect2] (styl) {Stylometric \\Attribution};
% \node [rect2, right of=styl, xshift=2cm] (stylmodel) {Linguistic Model \cite{uchendu2020authorship},
% Feature-based detector \cite{frohling2021feature}};

\node [rect3, below of = styl] (deep) {Deep learning-based Attribution};
\node [rect3, right of = deep, xshift=2cm, yshift=1.4cm] (deepglove) {Glove-based \\Attribution};

\node [rect3, below of = deepglove, yshift=0.4cm] (deepene) {Energy-based \\Attribution};

\node [rect3, below of = deepene, yshift=0.4cm] (deeptrans) {Transformer-based Attribution};

\node [rect4, below of=deep] (stat) {Statistical \\Attribution};  

\node [rect1, left of=stat, xshift=-2cm] (neural) {\textbf{Authorship \\Attribution \\for Neural Texts}};

\node [rect5, rounded corners, below of = stat] (hybrid) {Hybrid \\ Attribution};

\node [rect6, rounded corners, below of = hybrid] (human) {Human-based Evaluators};
\node [rect6, rounded corners, left of = human, xshift=5.6cm, yshift=0.8cm] (humantrain) {Human 
Evaluation without Training};
\node [rect6, rounded corners, below of = humantrain] (humanwtrain) {Human 
Evaluation \\with Training};

\path [line] (neural) |- (styl);
% \path [line] (styl) -- (stylmodel);

\path [line] (neural) |- (deep);
\path [line] (deep) -- (deepglove);
% \path [line] (deepglove) -- (deepglovemodel);

\path [line] (deep) -- (deepene);
% \path [line] (deepene) -- (deepenemodel);

\path [line] (deep) -- (deeptrans);
% \path [line] (deeptrans) -- (deeptransmodel);

\path [line] (neural) -- (stat);
% \path [line] (stat) -- (statmodel);

\path [line] (neural) |- (hybrid);
% \path [line] (hybrid) -- (hybridmodel);

\path [line] (neural) |- (human);
\path [line] (human) -- (humantrain);
\path [line] (human) -- (humanwtrain);

\end{tikzpicture}
\end{center}
        \caption{Taxonomy of Authorship Attribution models for NTD
        % \lee{"Neural Text Detectors" should be "Neural Authorship Attribution"? change "Classifers" to "Attribution". also, re-draw more condensely to fit a single column as currently, it takes too much space}
        }
    \label{fig:AATax}
\end{figure}
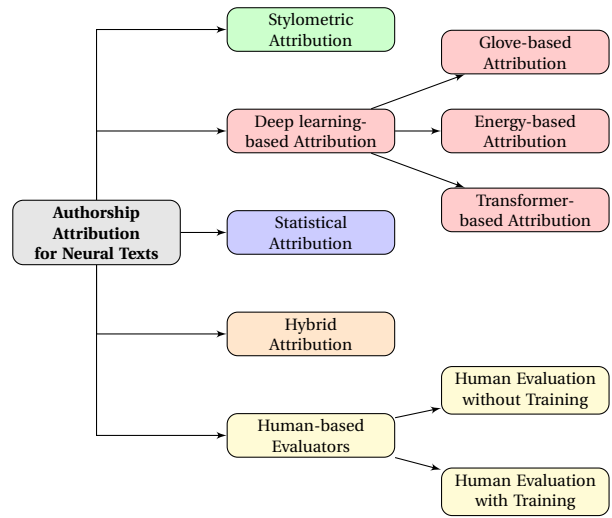

In the following, we survey recent AA solutions that are capable of handling neural texts in different ways, as illustrated in Figure \ref{fig:AATax}: 
%To solve these variations, researchers have proposed several AA techniques, which we categorized into:  
\textit{Stylometric Attribution},
\textit{Deep Learning-based Attribution}, 
\textit{Statistical Attribution}, and
\textit{Hybrid Attribution}. 
%See Figure \ref{fig:AATax} for a flowchart of the taxonomy of \textbf{Neural Authorship Attribution} models. \textbf{Neural Authorship Attribution} models are AA models for NTD. 
%We describe these categories in detail:

\subsection{Stylometric Attribution} \label{stylo}
Stylometry is the statistical analysis of the style of written texts. 
In traditional AA, stylometric classifiers 
are built using classical machine learning models trained on ensembles of  
style-based features such as $N$-grams, Part-of-Speech (POS), WritePrints \cite{abbasi2008writeprints}, 
LIWC (Linguistic Inquiry \& Word Count)\cite{pennebaker2001linguistic}, 
Readability score, and 
Empath \cite{fast2016empath}. 
% \lee{n-gram and POS are typical style-based features? anything else to put here?} 
This has been shown to be a successful approach for traditional AA tasks \cite{lagutina2019survey}.
% \thai{needs citation}. 
Due to such success, these models have been adopted and customized to the 
task of NTD.

The first attempt at a stylometric classifier to solve the 
AA task for $k>2$ authors is the \textit{Linguistic model} proposed by 
\cite{uchendu2020authorship}. 
It trains a Random Forest classifier with the Authorship Attribution-AA dataset 
in Table \ref{tab:data} and extracts an ensemble of 
stylometric features (e.g., \textit{entropy}, 
\textit{readability score}, \& 
\textit{LIWC} (Linguistic Inquiry \& Word Count) \cite{pennebaker2001linguistic}). 
The entropy feature counts the 
number of unique characters in the text. 
Readability scores represent the estimated educational 
level of the author of a piece of text based on  
lexicon usage. 
LIWC is a psycho-linguistic dictionary that counts the frequency of words 
that represents a psychological emotion or linguistic structure \cite{pennebaker2001linguistic}. 
% For instance, the words \textit{pal, buddy, coworker} count for 
% the social process, Friends. 
This \textit{Linguistic model}
achieves a 90\% F1 score 
and outperforms all the other deep learning-based models. 
However, this superior performance is a result of 
the small size of the dataset (only about 1k per data label) \cite{uchendu2020authorship}. 
Scaling up the data size in terms of labels and examples will make the AA task harder, 
and therefore cause the \textit{Linguistic model} to underperform. 
% \thai{this sentence is not very clear}. 
% However, \cite{uchendu2020authorship} claim that the high 
% performance of their \textbf{Linguistic model} 
% is due to insufficient examples of each of the data labels. 
This claim is confirmed in their second work using the TuringBench dataset
\cite{uchendu2021turingbench}. 
They compared SOTA deep-learning-based models (BERT and RoBERTa) with several 
stylometric classifiers - SVM (3-grams), 
WriteprintsRFC, Random Forest (w/ TF-IDF), Syntax-CNN, Ngram CNN, 
and N-gram LSTM-LSTM. RoBERTa outperforms all the stylometric classifiers
with about a 10-22\% increase in F1 scores. 

To further explore the benefits of stylometric features 
leveraged in the traditional AA community, \cite{frohling2021feature}
proposes a clever way to use them. This solution aims to solve the 
special case of AA, \textit{Turing Test} (TT). First, they identify different 
issues with NTGs which can be captured by
specific types of stylometric features. 
These issues in neural texts are categorized 
into 4 types: 
(1) \textit{Lack of syntactic and lexical diversity} which can be captured with 
Named Entity-tags, POS-tags, and neuralcoref extension\footnote{https://github.com/huggingface/neuralcoref}
(a tool for using a neural network 
to annotate and resolve coreference clusters) to detect coreference clusters; 
(2) \textit{Repetitiveness of words} which can be captured by collecting the number of
stop-words, unique words, and words
from “top-lists” of total words in a text.
Also, a ``conjunction overlap" measure is defined to calculate 
the overlap of the top-k words ($k=100, 1K, 10K$); 
(3) \textit{Lack of coherence} which can be captured using entity-grid 
representation to track
the appearance of the grammatical role of entities. 
They also use neuralcoref to detect 
coreference entity clusters;
(4) \textit{Lack of purpose} which is captured using a 
lexicon-package, empath \cite{fast2016empath} containing 200 linguistic features 
\cite{frohling2021feature}. 
% The results suggest that 
% some of the issue types are prevalent 
% in neural texts generated with top-k decoding strategy. 
% More quality-focused features (especially ones focused on 
% Lexical diversity) perform better than statistical 
% features such as the TF-IDF baseline. 
To evaluate the generalizability of these features, 
an ensemble of all the features is used to build a 
classifier - \textit{Feature-based detector}. 
This detector is trained and tested on different sizes of the GPT-2 
models. It is further evaluated on GPT-3 and GROVER texts. 
The classifier performs consistently in detecting texts 
generated by GPT-3, GROVER, and different model sizes of GPT-2,
suggesting it is generalizable to different NTG model sizes \cite{frohling2021feature}.
Further results suggest that some of the 4 categories of issues are prevalent in 
the top-k decoding strategy. Also, 
more quality-focused features (especially ones focused on 
Lexical diversity) perform better than statistical 
features such as the TF-IDF baseline.
% However, \cite{frohling2021feature} further claim that 
% while the model achieves high performance in the task, the lack 
% of diversity in the dataset (including discussion forums, and blog posts)
% fails to emulate a more realistic scenario. 

Scaling up and creating a more realistic scenario, 
\cite{diwan2021fingerprinting} collect 108 SubReddit blog posts generated by GPT-2
fine-tuned on 500K subreddit posts and comments. 
Every 108 labels indicate the 108 users of SubReddit (r/SubSimulatorGPT2). 
% Thus, this data is built for the 
% AA task with $k>2$ authors. 
These 108 authors are detected using a set of features called ``writeprints" features for the AA model
\cite{diwan2021fingerprinting}. Writeprints \cite{abbasi2008writeprints} is a
stylometric feature that collects lexical, content-based, 
and idiosyncratic features as the baseline. 
The writeprints classifier underperforms, 
compared to the RoBERTa-based baseline models. 
% The writeprints features underperform, 
% confirming \cite{uchendu2020authorship}'s claim that stylometric 
% features are insufficient for capturing the nuances in 
% linguistic patterns that are unique to all the NTGs
% vs. human. 
Similarly, a stylometric classifier with 791 stylometric features based on
\cite{kaur2019authorship} is used to detect neural texts. 
This classifier has 4 categories of features: 
\textit{Character, word, sentence,} and \textit{Lexical Diversity}
features. The classifier is an ensemble of classical ML models 
such as Random Forest and SVM and the stylometric features. 
BERT, a non-stylometric classifier, outperforms these stylometric classifiers 
significantly \cite{jones2022you}. 

Finally, stylometric classifiers are best used when the dataset size is small.
When data size increases, these models underperform, allowing deep learning-based models 
to outperform significantly.
Thus, we conclude that stylometric classifiers 
can only be considered good baselines since they underperform when the problem scales up. 
Another limitation of stylometry is that it fails
to detect neural misinformation due to NTG's capacity to generate 
consistent misinformation \cite{schuster2020limitations}.

% Feature-based detector, Linguistic detector, Fingerprint detector
\subsection{Deep Learning-based Attribution}
% \lee{too many terms below are in bold fonts, but they are not critical ones to highligh. Change them to italic fonts}

\textit{Stylometric} classifiers struggle to accurately assign 
the true authorship to human vs. neural texts. 
In Section \ref{stylo}, we observe that some of the stylometric 
classifiers were outperformed by deep learning-based models. 
Additionally, \cite{schuster2020limitations}'s findings of 
stylometry failing to detect neural misinformation, further 
calls for a different technique to solve 
the AA task for NTD. 
Therefore, researchers have adopted and advanced deep learning-based techniques for 
the attributing of neural vs. human text. 
% Deep learning techniques have 
% been shown to achieve close to human-level and sometimes
% above human-level performance. 
These models can be further categorized into 3 types of deep learning-based classifiers
- \textit{Glove-based, Energy-based,} and  \textit{Transformer-based} Attribution.
% To that end, there are 3 different 
% categories of deep learning models that have been used for 
% NTD. See below:

% Energy discriminator, GROVER detector, GPT-2 detector, OpenAI detector, 
% Looking-glass model
% RoBERTa-TT, BERT-TT, RoBERTa-Multi, BERT-Multi, 
% FAST 

\subsubsection{Glove-based Attribution} 
% Word2vec is an encoder-decoder word emdedding that captures 
% the co-occurrence statistics of words \cite{mikolov2013efficient}.
Glove is an unsupervised learning algorithm for extracting 
the representation of words. It aggregates global word-word 
\\co-occurrence statistics from a piece of text 
\cite{pennington2014glove}. 
% This algorithm was used to train several models of different sizes
% with Wikipedia, common crawl, and Twitter datasets. 
Using GloVe word embeddings with RNN and LSTM-based neural networks was 
considered SOTA until, 2018 (birth of BERT \cite{devlin2018bert}). 
Thus Glove-based classifiers now provide a good baseline for text classification tasks.
Some of the best-performing AA classifiers in the 
traditional AA communities are an
ensemble of stylometric features + GloVe pre-trained models w/
a neural network architecture. 
Several Glove-based classifiers have been used as baselines for the 
NTD task. 
\textit{Syntax-CNN} \cite{zhang2018syntax},
\textit{N-gram CNN} \cite{shrestha2017convolutional}, and 
\textit{N-gram LSTM-LSTM} \cite{jafariakinabad2019syntactic} 
are baselines for \cite{uchendu2021turingbench}. 
\textit{Embedding}, 
\textit{RNN}, 
\textit{Stacked-CNN}, 
\textit{Parallel-CNN}, and 
\textit{CNN-RNN}, are baselines for
\cite{uchendu2020authorship}. 
They demonstrated that 
Glove-based classifiers are unsuitable for 
solving the AA problem when there are $k>2$ authors.  
% This further confirms that the AA task with $k>2$ authors
% is the most non-trivial of the AA task variations. 
Furthermore, the \textit{Fingerprint detector} is compared with 4 baseline models - 
\textit{Gaussian Naive Bayes},
\textit{Random Forest},
\textit{Multi-layer Perceptron}, and
\textit{CNN} classifiers 
using the Glove word embeddings of the data. 
They underperform significantly 
\cite{diwan2021fingerprinting}. 

Lastly, 
\textit{Sentiment modeling detector}, a variation of 
sentiment neuron used to learn
a single-layer multiplicative LSTM (mLSTM) \cite{krause2016multiplicative}
is used to detect texts generated by GPT-2 \cite{adelani2020generating}. 
The goal is to force the model to focus on a specified sentiment
\cite{adelani2020generating}. This model
outperforms and sometimes performs comparably with the baseline -
original mLSTM model \cite{krause2017multiplicative}, achieving a 70\% accuracy in 
detecting GPT-2 generated Amazon and Yelp reviews \cite{adelani2020generating}.

\subsubsection{Energy-based Attribution} \label{energy}
% Energy discriminator,
Energy-based models (EBMs) are un-normalized generative 
models based on energy functions \cite{lecun2006tutorial}. 
Using the energy functions, EBMs model the 
probability distribution of its training data and generates high 
quality data similar to the training set 
\cite{lecun2006tutorial}. 
It is also able to adapt to changes in the Language model. 
Due to this capability, \textit{Energy-based classifier} 
is proposed by \cite{bakhtin2019real} to detect neural texts. 
This classifier is trained on 3 datasets of different domains - 
\textit{Books}, \textit{CCNews}, and \textit{Wikitext}. 
Three sizes of the GPT-2 model are used for the generator architectures
and three architectures are used for the energy function - 
\textit{Linear}, \textit{BiLSTM}, and \textit{Transformer}. 
Their findings suggest that: 
(1) as the NTG increases in size, 
the harder the AA task becomes;
(2) the biggest energy function 
(i.e.\textit{Transformer}) performs the best in detecting 
texts generated from large language models (e.g., GPT-2 Large \& XL); 
(3) as the length of texts increases, the task becomes even more non-trivial; and
(4) the classifiers are able to generalize to 
data that it is not trained on. 

In addition, 
EBMs are very expensive to train and do not scale well \cite{bakhtin2019real}.
While the \textit{Energy-based classifier} performs well in the 
AA problem, achieving over a $90\%$ in all experiments, 
applying the classifier to a much larger dataset is too 
expensive to justify.

\subsubsection{Transformer-based Attribution}
Since the advent of the Transformer architecture, 
the current SOTA text classification models are Transformer-based 
models. Based on Section \ref{energy}, we observe that 
large models are better at detecting neural texts. 
However, since EBMs are too
expensive, several researchers have adopted Transformer-based 
models (i.e., BERT, RoBERTa, etc.) for the AA tasks. 
In Section \ref{stylo}, most of the stylometric 
classifiers were outperformed by Transformer-based classifiers. This further 
supports the application of this classification technique to the AA problem. 
 
\textit{GROVER detector}\footnote{https://grover.allenai.org/detect} \cite{zellers2019defending}
is trained on texts generated by the GROVER NTG. 
% To be socially responsible, GROVER \cite{zellers2019defending} creators 
% built a detector to accompany their NTG - GROVER. 
% It has an online demo\footnote{https://grover.allenai.org/detect}.
It is built with similar architecture as the GPT-2 classifier.
\textit{GROVER detector} has been evaluated on neural texts 
generated by different NTGs (GPT-2, FAIR, PPLM, etc.) \cite{uchendu2020authorship, uchendu2021turingbench, zhong2020neural,gagiano2021robustness}. 
It performs well at detecting neural texts generated 
by older NTGs (2018-2019), however, struggles at detecting more recent NTGs
accurately. For instance, \textit{GROVER detector} achieved a 58\% F1 score 
in detecting GPT-3 texts with the TuringBench dataset \cite{uchendu2021turingbench}.
% \textit{GROVER detector} makes a good 
% baseline for detecting more recent neural texts too. 
Next, GPT-2 has a detector trained to detect texts generated by GPT-2 
- \textit{GPT-2 detector}\footnote{https://huggingface.co/openai-detector/}  \cite{gpt2outputdetector}. 
% It also has a demo\footnote{https://huggingface.co/openai-detector/} 
% released by Huggingface. 
Just like \textit{GROVER detector}, 
\textit{GPT-2 detector} has also been evaluated on neural texts generated 
by different NTGs \cite{uchendu2020authorship,uchendu2021turingbench,wolff2020attacking, gao2022comparing} and more easily detects older NTGs 
than newer NTGs. 
This is confirmed with \textit{GPT-2 detector}'s performance in 
detecting GPT-3 texts, achieving a 53\% F1 score \cite{uchendu2021turingbench}. 
The reason is that newer NTGs, such as GPT-3 tend to generate more human-like texts which confuse SOTA older AA models, 
like \textit{GPT-2 detector}.
% \thai{This happens because...}

% \thai{This paragraph is verbose and should be shorten to half (like for XLNet-FT).}
There are two RoBERTa-based models 
(base \& large) trained on GPT-2 dataset\footnote{https://github.com/openai/gpt-2-output-dataset} 
in huggingface repo\footnote{https://huggingface.co/roberta-base-openai-detector},\footnote{https://huggingface.co/roberta-large-openai-detector}. 
We call both the base and large models, 
\textit{OpenAI  detector}. 
This AA model has been evaluated on neural texts generated by 
different NTGs \cite{wolff2020attacking, uchendu2021turingbench}. 
\textit{OpenAI  detector} is the same model as \textit{GPT-2 detector},
except that \textit{OpenAI detector} has been re-purposed for 
the AA multi-class setting, while \textit{GPT-2 detector} remains 
for the AA binary setting. 
% \cite{uchendu2021turingbench} trained  
% \textit{OpenAI  detector} on the TuringBench dataset\footnote{https://huggingface.co/datasets/turingbench/TuringBench} 
% to perform the AA task. 
\textit{OpenAI  detector} performs comparably to the 
AA models - \textit{BERT-Multi} and \textit{RoBERTa-Multi} when evaluated on 
the TuringBench-AA dataset
\cite{uchendu2021turingbench}. 
\textit{BERT-Multi} and \textit{RoBERTa-Multi} are BERT and RoBERTa base
models, respectively trained on the TuringBench-AA dataset.
\textit{BERT-TT} and 
\textit{RoBERTa-TT} outperform 
\textit{GROVER detector} and \textit{GPT-2 detector} 
when evaluated on the TuringBench-TT
dataset \cite{uchendu2021turingbench}. 
\textit{BERT-TT} and \textit{RoBERTa-TT} are BERT and RoBERTa base
models, respectively trained on the TuringBench-TT dataset. \textit{BERT-TT}, 
outperforms all the models, including \textit{RoBERTa-TT} significantly. 
% Thus, the authors claim that the 
% insufficient training data may have caused \textbf{RoBERTa-TT}
% to be unstable which made it underperform. 
% However, while all 
% these model perform reasonably well, \textbf{BERT-TT} and \textbf{RoBERTa-Multi} 
% outperforms other models, achieving an 88\% and 81\% F1 score, 
% respectively. 
Furthermore, for all 19 pairs of human vs. NTG, 
no model consistently outperforms all other 
models. In fact, \textit{GROVER detector}
and \textit{GPT-2 detector} performs poorly in detecting 
texts generated by GROVER and GPT-2, respectively \cite{uchendu2021turingbench}. 

\textit{XLNet-FT} is a fine-tuned XLNet 
classification model trained to 
detect texts generated by GPT-2 \cite{munir2021through}. 
The generalizability of the model is evaluated on 
different subreddit post domains.
\textit{XLNet-FT} performs consistently, achieving over a 
90\% accuracy in all experiments, suggesting that 
it is generalizable \cite{munir2021through}. 
However, when \textit{XLNet-FT} is further evaluated 
on neural texts generated by top-p and top-k
decoding strategy, 
there is a significant drop in accuracy, 
suggesting that the AA model may not be generalizable. 
% We need neural text detectors that can detect 
% all neural texts no matter the decoding strategies used for 
% generation.
% They also perform very poorly in detecting XLM and XLNET texts, possibly because their generations are of very bad quality \cite{stiff2022detecting}. 
% Furthermore, a common phenomenon in all AA models evaluated is that they underperforms in 
% detecting social media posts, 
% especially for short posts like tweets rather than news articles. 

Using an \textit{in-the-wild} dataset concept, 
\textit{RoBERTa-Defense} is evaluated on 4 types of 
\textit{in-the-wild} datasets \cite{pudeepfake}.
\textit{In-the-wild} datasets are test sets generated from an entirely different NTG from the training set.
\textit{RoBERTa-Defense} is trained on human \& GROVER Real news in Table \ref{tab:data} 
and compared to 
other SOTA AA models - \textit{GLTR} (with two different LMs - BERT \& GPT-2 which results in \textit{GLTR-BERT} \& \textit{GLTR-GPT2}),  
\textit{GROVER detector}, \textit{BERT-Defense}, and \textit{FAST}.
\textit{RoBERTa-Defense}
outperforms all other models significantly. 
\textit{BERT-Defense}, a BERT model fine-tuned on GPT-2 Large Webtext 
dataset in Table \ref{tab:data} from which \textit{RoBERTa-Defense} is inspired
is evaluated with different decoding strategies \cite{ippolito2020automatic}. 
\textit{BERT-Defense} is trained and tested on neural texts 
generated by different decoding strategies - top-k, top-p, untruncated random, and mixed (i.e., dataset containing equal amounts of each strategy) \cite{ippolito2020automatic}. 
The classifier trained on the mixed dataset is the most generalizable classifier. Similarly, \textit{RoBERTa,
BERT}, \textit{ELECTRA} \cite{clark2020electra}, and 
\textit{ALBERT} \cite{lan2019albert}
are evaluated on \textit{in-the-wild} dataset \cite{pagnoni2022threat}. These models are evaluated, 
specifically on 
out-of-domain COVID-19 human-written vs. neural news. 
The neural news is generated with GPT-2 small, medium, 
large, XL, and GPT-Neo using top-p and top-k decoding strategies \cite{pagnoni2022threat}. \textit{ELECTRA} performs better at 
generalizing to out-of-domain neural texts, achieving an 
average accuracy of 86\% on all out-of-domain datasets.

% \cite{ippolito2020automatic} proposes 
% \textbf{BERT-Defense} which is BERT fine-tuned on GPT-2 Large Webtext 
% dataset in Table \ref{tab:data}. 
% Generalizability of \textbf{BERT-Defense} is evaluated 
% with decoding strategy. 
% \textbf{BERT-Defense} is trained and tested on neural texts 
% generated by different decoding strategies - top-k, top-p, untruncated random, and mixed (i.e., dataset containing equal amounts of each strategy) \cite{ippolito2020automatic}. 
% They find that training with the mixed dataset performs well in detecting 
% each of the other strategies.  

Due to the nuances of neural texts, \cite{constrabert} 
proposes to combine the advantages of contrastive learning \cite{gao2021simcse} with 
a Transformer-based classifier. 
Thus, they propose \textit{Constra-DeBERTa} which is a
DeBERTa model \cite{he2021deberta} trained with a contrastive learning 
approach. 
Contrastive learning is a technique that clusters similar examples
together and separates dissimilar examples in a representation space
\cite{constrabert,gao2021simcse}.
However, while 
\textit{Constra-DeBERTa}
outperforms other SOTA traditional AA models, it only performs 
comparably to \textit{RoBERTa-Multi} on detecting the 
TuringBench dataset. 
Similarly, \cite{rosati2022synscipass} trains \textit{DeBERTa v3} \cite{he2021debertav3} on the 
SynSciPass dataset to answer the question, if a piece of text is neurally generated, 
how was it generated? The answer choices are \textit{generated, paraphrased}, or 
\textit{translated} vs. human-written. Using this dataset, 
\textit{DeBERTa v3}
achieves a 99.6\% F1 score.
Next, \textit{DistilBERT-Academia} is trained on 
human vs. GPT-2 academic abstracts and papers \cite{liyanage2022benchmark} and 
achieves a 62.5\% and 70.2\% accuracy on the FULL and PARTIAL
Academic datasets, respectively. 
Furthermore, \textit{RoBERTa w/ GCN} (Graph Convolutional Networks) is used to detect human-written news with entities manipulated 
and replaced by GPT-2 \cite{jawahar2022automatic}. 
The GCN \cite{kipf2016semi} model is used to capture factual knowledge of neural vs. human news articles. RoBERTa outperformed the proposed model - \textit{RoBERTa w/ GCN} on most of the GPT-2 detection tasks \cite{jawahar2022automatic}.

Lastly,  \textit{ruBERT}, a Russian BERT model is used  to distinguish Russian neural 
texts from Russian human-written texts as a shared task \cite{shamardina2022findings}. 
This fine-tuned \textit{ruBERT} (Russian BERT)
achieves 82.6\% accuracy for $k=2$ authors and 64.5\% accuracy for $k>2$ authors
\cite{automik22}. 
% Next, fine-tuning RoBERTa on TweepFake (human vs. bot tweets) achieves
% a 90\% accuracy on detecting machine-generated tweets \cite{fagni2021tweepfake}. 
Finally, using an \textit{Ensemble} classifier (BART, BERTweet, and TwitterRoberta), \cite{gambini2022pushing}
achieves an 84\% accuracy in distinguishing between GPT-2 and human tweets. 
However, using the same model for GPT-3 generated tweets achieves a 
54\% accuracy \cite{gambini2022pushing}. This suggests that GPT-3 generates 
more human-like tweets than GPT-2.

% To be specific, the per class F1 scores are 97.4\%, 96.9\, 99.8\%, and 96.2\% for generation, paraphrase, human-written,
% and translation classes respectively \cite{rosati2022synscipass}.
% % \cite{automik22} reports are part of RuATD (Russian Artificial Text Detection) 
% % shared task \cite{shamardina2022findings}. 

% GROVER detector, GPT-2 detector, OpenAI detector,
% Looking-glass model
% RoBERTa-TT, BERT-TT, RoBERTa-Multi, BERT-Multi, 
% FAST 

\begin{figure}[t]
    \centering
    \includegraphics[width=0.5 \textwidth]{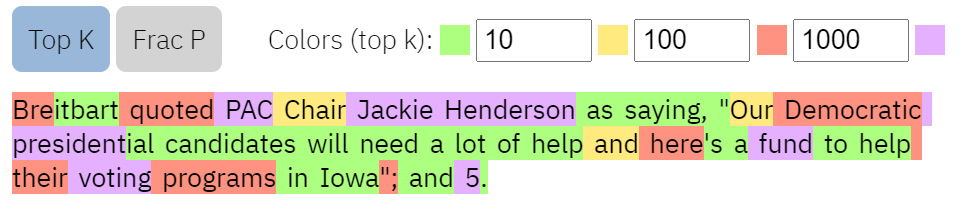}
    \caption{\cite{uchendu2021turingbench} used GLTR \cite{gehrmann2019gltr} on 
    GPT-3 texts. 
    \textcolor{green}{Green} represents the most probable words (top-10); \textcolor{yellow}{yellow} the 2nd most probable (next top-100 probable words);
    \textcolor{red}{Red} the least probable (next top-1000 probable words); 
    and \textcolor{violet}{purple} the highest improbable words.
    The hypothesis is that neural texts are often populated with mostly \textcolor{green}{Green} and \textcolor{yellow}{yellow} words. However, we see that texts generated by GPT-3 are very human-like according to the hypothesis.
    }
    \label{fig:gltr}
\end{figure}

\subsection{Statistical Attribution}
We observe that while there are some well-performing 
stylometric and deep learning-based models, 
there is still a lot of room for improvement, especially 
in building generalizable models. The biggest feat is in 
building classifiers that perform consistently well in detecting 
neural texts generated by top-p and top-k decoding strategies. 
Thus, statistical models are proposed to combat 
these limitations. 
To assess the validity of statistical techniques, 
a hypothesis using $k$-order
Markov approximations are formulated \cite{varshney2020limits}. 
This statistical formulation proves the hypothesis that 
human language is stationary and ergodic as opposed to neural 
language. The formal hypothesis testing framework is used to 
establish limits in error exponents between human and neural text 
\cite{varshney2020limits}. This suggests that statistical 
AA models for neural texts could be successful.
There are currently only four 
statistical classifiers that capture the writing style of neural texts by modeling their statistical distribution. 

The first statistical AA classifier proposed for NTD is 
\textit{GLTR} \cite{gehrmann2019gltr}. \textit{GLTR} performs 
3 tests - 
(1) probability of the word; 
(2) the absolute rank of the word; 
(3) the entropy of the predicted distribution to detect
neural texts. \textit{GLTR} has a demo\footnote{http://gltr.io/dist/index.html} that highlights 
words by distribution and 
is used to assist humans in detecting neural texts. 
See Figure \ref{fig:gltr} \cite{uchendu2021turingbench}
to see how \textit{GLTR} detects texts generated by GPT-3. 
% Green represents the
% top-10 probable words; yellow the next top-100 probable;
% Red the next top-1000 probable words; and purple the highly improbable words. The hypothesis is that neural texts are often populated with mostly Green and yellow words.
This classifier improved human performance in detecting neural texts 
from 54\% to 72\%. 
However, since 2019 when it was built, more 
sophisticated NTGs have been 
built. These newer NTGs have more human-like statistical distribution, 
making it harder for \textit{GLTR} to distinguish neural texts from human texts.
\textit{GLTR}, especially underperforms in detecting GPT-3 texts, 
achieving a 35\% F1 score, which is significantly less than a random guess (50\%)
\cite{uchendu2021turingbench}. 

\textit{MAUVE}  is another statistical classifier
\cite{pillutla2021information}. 
This AA classifier measures 
the gap between the distribution of human and 
neural texts. Using KL-divergence, \textit{MAUVE} models two types 
of errors that highlight the unique distributions in human 
vs. neural texts \cite{pillutla2021information}. Human detection of texts generated by GROVER and GPT-2 correlated strongly with \textit{MAUVE}'s highlight of differences between human and neural texts. 
\textit{Distribution detector},
an unsupervised AA model for
calculating the distribution of repeated n-grams in neural texts is 
used to detect neural texts \cite{galle2021unsupervised}. 
The hypothesis is that NTGs are more repetitive
than humans which is also one of the hypotheses of \cite{frohling2021feature}. 
\textit{Distribution detector}
achieves over 90\% and 80\% accuracy in detecting texts generated by GPT-2 using 
top-k and top-p decoding strategies, respectively. 

Most recently,  a zero-shot unsupervised neural text detector, \textit{DetectGPT}~\cite{mitchell2023detectgpt}, is proposed. The hypothesis of this statistics-based detector 
is that neural texts tend to lie in areas of negative curvature of the 
log probability function \cite{mitchell2023detectgpt}. 
Therefore, if a piece of neural text is perturbed, the curvature of the log probability 
will still bear a strong similarity to the unperturbed neural texts. 
Hence, \cite{mitchell2023detectgpt} considers an AO technique that slightly modifies 
the original neural text, while preserving semantics. After running several perturbation experiments, a threshold for perturbation discrepancy is defined and used to detect neural text. Thus, by measuring the curvature of log probability with the strict constraint of perturbation discrepancy threshold, 
\textit{DetectGPT} can detect texts generated by a 
neural method. Finally, \textit{DetectGPT} detects GPT-3 generated texts with an average 
of 85\% AU-ROC, performing comparably to RoBERTa \cite{liu2022coco}.
Lastly, \textit{DetectGPT} has an online demo\footnote{https://detectgpt.ericmitchell.ai/}.

% GLTR, MAUVE, Distribution detector, DetectGPT

\subsection{Hybrid Attribution}
There are advantages in each of the AA model categories,
however, each of them is still unable to accurately 
attribute neural vs. human texts to their authors consistently. 
Furthermore, the issue of different decoding strategies, 
also, make the AA models unable to generalize well 
\cite{holtzman2019curious,pudeepfake,frohling2021feature,ippolito2020automatic}. 
Therefore, a few researchers have proposed hybrid classifiers, 
which are ensembles of two or more of the AA categories. 

\textit{TDA-based detector}, 
an ensemble of the Transformer-based and statistical AA techniques
is used to solve the NTD task. 
% \cite{kushnareva2021artificial} proposed a novel approach for 
% neural text detection 
% using Topological Data Analysis (TDA) techniques - \textbf{TDA-based detector}. 
% This is an ensemble of Transformer-based and statistical features. 
This classifier involves 
obtaining the attention matrices of 
BERT's word representations of texts generated by GPT-2 and GROVER. 
Next, using these BERT word representations,
3 interpretable TDA-based features are extracted: 
(1) \textit{Topological Features}: Calculating the first 2 betti 
numbers (i.e., topological features based on the connectivity of n-dimensional simplicial complexes) of the attention matrices; 
(2) \textit{Features derived from barcodes}: Calculating characteristics of 
the barcode plots of the persistent homology 
of the attention matrix; 
(3) \textit{Features based on the distance to patterns}: Calculating the 
distance in features in the attention graph. This feature is used
to capture linguistic patterns. 
Finally, \textit{TDA-based detector} is a 
logistic regression model, trained on 
an ensemble of the three TDA features \cite{kushnareva2021artificial}. 
Comparing this model to pre-trained and fine-tuned BERT models, 
it performs comparably to BERT models fine-tuned on GPT-2 small Webtext, 
GPT-2 XL Amazon Reviews, and GROVER News \cite{kushnareva2021artificial}. 
While more analysis is required to 
understand why the \textit{TDA-based detector} performs well, 
this approach has interpretable qualities that  
should be explored in future work. 

\textit{Fingerprint detector} is another hybrid classifier 
for NTD. 
It is 
an ensemble of fine-tuned RoBERTa
embeddings and CNN classifier \cite{diwan2021fingerprinting}. 
\textit{Fingerprint detector} solves the AA task by 
detecting 108 neural authors.
The \textit{Fingerprint detector} achieves a 70\% accuracy (top-10).
This shows promise in the area of detection of neural texts \textit{in-the-wild},
where there are $k>100$ authors. 
To continue the quest for generalizable classifiers, 
\textit{FAST} uses a Graph 
Neural Network (GNN) architecture with RoBERTa to 
capture the factual 
structure of neural and human texts \cite{zhong2020neural}.  
It detects neural texts by calculating the RoBERTa word embeddings of 
the texts and then extracting the graphical representation 
\cite{zhong2020neural}. Next, it uses a GNN to capture 
sentence representations that consider coherence \cite{zhong2020neural}. 
The experiments included detecting texts generated by GROVER and 
GPT-2. \textit{FAST} outperforms \textit{GROVER detector} and other
baselines significantly. 
% for both the P2 and P4 AA tasks. 
Surprisingly, it performs the best at detecting 
human-neural text pairs, 
achieving over 93\% accuracy while unpaired texts achieve over 84\% accuracy. 

\textit{CoCo} is a coherence-based contrastive learning model \cite{liu2022coco}. 
It is architecturally similar to \textit{FAST} in that it uses a graphical neural network 
to represent the sentences of human-written vs. neural texts. 
Since human-written texts are more coherent than neural texts, they
sentences share more entities \cite{liu2022coco}. 
The texts are represented as RoBERTa embedding weights which are concatenated with the 
sentence-level graphical representations of the texts. These 
concatenated features are input for an LSTM with attention. Lastly, 
\textit{CoCo} is trained using the
sum of the coss-entropy loss and contrastive loss \cite{liu2022coco} to improve model performance. 
Thus, it achieves an F1 score of 83\% and 94\% using the full dataset for 
GROVER, and GPT-2, respectively \cite{liu2022coco}. Furthermore, 
calculating the graphical metrics showed that in terms of the number of vertex and edges, human-written texts have significantly more graphical features than neural texts.

% \thai{Since this section involves hybrid modality (not only texts, as defined in AA). I would suggest make this as a separate open-problem in Section 6. That would be more clear, since section 
% 3 categories different models based on how they extract the features, not on the input modality.} 
Lastly, \cite{tan2020detecting} explore 
the most realistic scenario of misinformation 
where malicious users of NTGs, pair neurally generated 
misinformation with fake/real images to increase 
the authenticity of the news article. 
\textit{DIDAN} is a multi-modal NTD evaluated on a multi-modal 
dataset containing 
both texts and images \cite{tan2020detecting}. 
% \cite{tan2020detecting} studies a more realistic scenario 
% where the neurally generated news is paired with fake images to 
% increase the authenticity of the news article. 
% To detect this cross-modal neural news dataset, \cite{tan2020detecting}, 
% proposes an ensemble-based model, \textit{DIDAN}. 
This NTD encodes the texts with BERT encoder and  
investigates Visual-Semantic representations from images and texts. These features are used 
to evaluate the semantic consistency between linguistic and visual components in a news article 
\cite{tan2020detecting}. 
An \textit{authenticity score} is defined to represent the probability of an article being human-written. 
It is calculated by extracting the 
co-occurrences of named entities in the news articles and captions
\cite{tan2020detecting}. 
They build different variations of dataset, some only 
containing text. Using only the text dataset, 
\textit{DIDAN} is compared to \textit{GROVER detector} as well as 
other baseline models, and outperforms all of them
\cite{tan2020detecting}.

% and Canonical Correlation
% Analysis (CCA) which extracts the shared semantic
% space between two sets of paired features \cite{tan2020detecting}. 

% DL + ML
% TDA
% Fingerprint
% DIDAN

\begin{table*}[]
    \centering
    \resizebox{13cm}{!}{
    \begin{tabular}{|c|c|}
    \hline
     \textbf{Authorship Obfuscation}  & \textbf{Adversarial Attack}  \\
     \hline
     
      Has a strict definition of writing style   &  
      Has a loose definition of writing style \\
      
      \hline
      
      Requires consistent/uniform change in writing style &
      Does not require consistent/uniform change in writing style \\
      
      \hline 
      
      Requires semantics to be preserved &
      Does not always require semantics to be preserved \\
      
      \hline 
      % The ultimate goal is to evade human and machine detection &
      % The ultimate goal is to evade machine detection \\
      % \hline

    \end{tabular}
    }
    \caption{Differences between Authorship Obfuscation and Adversarial Attack}
    \label{tab:ao}
    % \vspace{-10pt}
\end{table*}

% \begin{figure}
%     \centering
%     \includegraphics[width=0.65\linewidth]{figures/AO.drawio.png}
%     \caption{Taxonomy of Authorship Obfuscation techniques for obfuscating the Linguistic structures of neural texts}
%     \label{fig:aotax}
% \end{figure}

\tikzstyle{rect} = [draw, rectangle, rounded corners, fill=red!20, 
text width=6em, text centered, minimum width=2em]
\tikzstyle{rect1} = [draw, rectangle, rounded corners, fill=gray!20, text width=10em, text centered, minimum width=2em]
\tikzstyle{rect2} = [draw, rectangle, rounded corners, fill=green!20, text width=10em, text centered, minimum width=10em]
\tikzstyle{rect3} = [draw, rectangle, rounded corners, fill=red!20, text width=10em, text centered, minimum width=10em]
\tikzstyle{rect4} = [draw, rectangle, rounded corners, fill=blue!20, text width=10em, text centered, minimum width=10em]
\tikzstyle{rect5} = [draw, rectangle, rounded corners, fill=orange!20, text width=10em, text centered, minimum width=10em]
\tikzstyle{rect6} = [draw, rectangle, rounded corners, fill=yellow!20, text width=10em, text centered, minimum width=10em]

\tikzstyle{elli} = [draw, ellipse, fill=white!20,
minimum height=2em]
\tikzstyle{circ} = [draw, circle, fill=white!20, minimum width=8pt,
inner sep=10pt]
\tikzstyle{diam} = [draw, diamond, fill=white=!20, text width=6em,
text badly centered, inner sep=0pt]
\tikzstyle{line} = [draw, -latex']

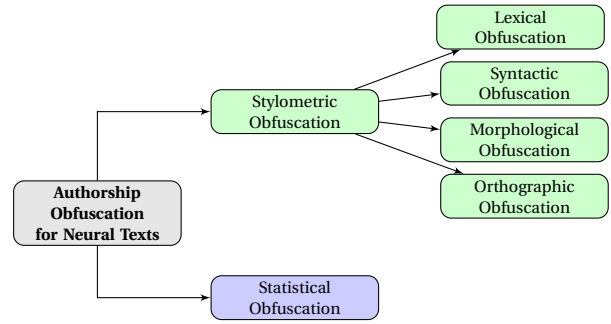
\begin{figure}
\begin{center}
\begin{tikzpicture}[node distance = 1.5cm, scale=0.75, transform shape]
\node [rect2] (styl) {Stylometric \\Obfuscation};

\node [rect2, right of = styl, xshift=2.5cm, yshift=1.5cm] (lexical) {Lexical \\Obfuscation};
\node [rect2, below of = lexical, xshift=0.07cm, yshift=0.5cm] (syntax) {Syntactic \\Obfuscation};
\node [rect2, below of = syntax,  yshift=0.5cm] (morpho) {Morphological Obfuscation};
\node [rect2, below of = morpho,  yshift=0.5cm] (ortho) {Orthographic Obfuscation};

% \node [rect3, below of = styl] (deep) {Deep Learning Obfuscation};
\node [rect1, left of=deep, xshift=-2cm] (neural) {\textbf{Authorship \\Obfuscation \\for Neural Texts}};

\node [rect4, below of=deep] (stat) {Statistical \\Obfuscation};

% \node [rect5, rounded corners, below of = stat] (hybrid) {Hybrid Classifiers};

% \node [rect6, rounded corners, below of = hybrid] (human) {Human-based Evaluators};

\path [line] (neural) |- (styl);
\path [line] (styl) -- (lexical);
\path [line] (styl) -- (syntax);
\path [line] (styl) -- (morpho);
\path [line] (styl) -- (ortho);

% \path [line] (neural) -- (deep);

\path [line] (neural) |- (stat);
\end{tikzpicture}
\end{center}
        \caption{Taxonomy of Authorship Obfuscation algorithms/techniques for NTD
        }
    \label{fig:aotax}
    % \vspace{-15pt}
\end{figure}

\begin{table*}[t!]
% \footnotesize
\centering
\resizebox{14cm}{!}{
\begin{tabular}{|c|c|c|c|c|c|}
\hline
        \textbf{AO technique} & \textbf{Scenario} &   \textbf{Category}  & \textbf{Interpretable} & 
        \begin{tabular}[c]{@{}c@{}}
        \textbf{Preserves} \\ \textbf{semantics} 
        \end{tabular}
        & 
        \textbf{Obfuscated dataset} \\
       
        \hline
       Homoglyph \cite{wolff2020attacking,gagiano2021robustness} & Black-box & 
       Stylometric (Orthographic)  
       & \cmark & \cmark & GROVER \& GPT-2  \\
       \hline 
       Upper/Lower Flip \cite{gagiano2021robustness} &  Black-box & 
       Stylometric (Morphological) 
       & \cmark & \cmark & GROVER \& GPT-2  \\
       \hline 
       DeepWordBug   \cite{crothers2022adversarial,stiff2022detecting} & Black-box & 
       Stylometric (Lexical) 
       &  &  & GPT-2 \& GPT-3\\
       \hline
       Misspellings attack \cite{wolff2020attacking, gagiano2021robustness} & Black-box & 
        Stylometric (Lexical) 
       & \cmark & &  GROVER \& GPT-2 \\
       \hline 
       Whitespace attack \cite{gagiano2021robustness} & Black-box & Stylometric (Lexical) 
       & \cmark & &  GROVER \& GPT-2 \\
       \hline
       Deduplicate tokens \cite{puunraveling} & Black-box &  Stylometric (Lexical) 
       & \cmark & & Human-Machine Pairs \\
       \hline 
       Shuffle tokens \cite{puunraveling} & Black-box &  Stylometric (Syntactic) 
       & \cmark &  & Human-Machine Pairs  \\
       \hline 
        \begin{tabular}[c]{@{}c@{}}
       Retain only (non)-stopwords \cite{puunraveling} 
       \end{tabular}
       & Black-box &  Stylometric (Syntactic) 
       & \cmark &  & Human-Machine Pairs \\
       \hline
       \begin{tabular}[c]{@{}c@{}}
       Retain tokens in \\high/low frequency \cite{puunraveling} 
       \end{tabular}
       &  White-box &  Statistical 
       &  \cmark &  & Human-Machine Pairs \\
       \hline 
    \begin{tabular}[c]{@{}c@{}}
       Replace text with \\likelihood ranks \cite{puunraveling} 
       \end{tabular}
       &  White-box &   Statistical 
       & \cmark &   & Human-Machine Pairs  \\
       \hline 
       \begin{tabular}[c]{@{}c@{}}
       Replace text with specific\\ linguistic features \cite{puunraveling} 
       \end{tabular}
       & White-box &   Statistical
       & \cmark &  & Human-Machine Pairs \\
       \hline 
       TextFooler \cite{crothers2022adversarial,pudeepfake} & Black-box & Stylometric (Lexical) &  &  \cmark & 
       \begin{tabular}[c]{@{}c@{}}
       GPT-2 medium, GPT-2 \\XL, GPT-3, GROVER 
       \end{tabular}
       \\
       \hline 
       Varying sentiment \cite{bhat2020effectively} & Black-box & Stylometric (Lexical) & \cmark &  &
       GROVER \\
       \hline 
       Source-target exchange \cite{bhat2020effectively} & Black-box & Stylometric (Syntactic) & \cmark &  &
       GROVER \\
       \hline
       Entity replacement \cite{bhat2020effectively} & Black-box &  Stylometric (Lexical) & \cmark &   &
       GROVER \\
        \hline
       Alter numerical facts \cite{bhat2020effectively} & Black-box &  Stylometric (Syntactic)
       & \cmark &  & GROVER \\
       \hline
       Syntactic perturbations \cite{bhat2020effectively} & Black-box & Stylometric (Syntactic)
       & \cmark &  & GROVER  \\
       \hline
       Article shuffling \cite{bhat2020effectively} & Black-box &  Stylometric (Syntactic)
       & \cmark &  & GROVER \\
       \hline
       \begin{tabular}[c]{@{}c@{}}
       Selecting highest human\\-class probability \cite{automik22}
       \end{tabular} 
       & Black-box &  Statistical &
       &  & \begin{tabular}[c]{@{}c@{}}
       Russian neural \& \\ human-written texts 
       \end{tabular}  \\
       \hline 
       \begin{tabular}[c]{@{}c@{}}
       Adding detector's log-probability\\ to sampling technique \cite{automik22} 
       \end{tabular}
       & White-box &  Statistical &  &
       &  
       \begin{tabular}[c]{@{}c@{}}
       Russian neural \& \\ human-written texts 
       \end{tabular}
       \\
       \hline
       \begin{tabular}[c]{@{}c@{}}
       Varying the text decoding \\strategy (and its parameters) \cite{pudeepfake}
       \end{tabular}
       & White-box &  Statistical &  &
       \cmark &
       \begin{tabular}[c]{@{}c@{}}
       GPT-2 Large, GPT-2 XL, \\GPT-3, GROVER 
       \end{tabular}
       \\
       \hline 
       \begin{tabular}[c]{@{}c@{}}
       Varying the number of \\priming tokens \cite{pudeepfake} 
       \end{tabular}
       & White-box &  Statistical &  &  &
       \begin{tabular}[c]{@{}c@{}}
       GPT-2 Large, GPT-2 XL, \\GPT-3, GROVER 
       \end{tabular} \\
       \hline 
       DFTFooler \cite{pudeepfake} & Black-box & Stylometric (Lexical)  &  & \cmark &
       \begin{tabular}[c]{@{}c@{}}
       GPT-2 Large, GPT-2 XL, \\GPT-3, GROVER 
       \end{tabular} \\
       \hline 
       Random Perturbations \cite{pudeepfake} & Black-box & Stylometric (Lexical) &  &  &
       \begin{tabular}[c]{@{}c@{}}
       GPT-2 Large, GPT-2 XL, \\GPT-3, GROVER 
       \end{tabular} \\
       \hline 
       ALISON \cite{alison} & Black-box &  Stylometric (Syntactic) &  \cmark  & \cmark &
       TuringBench  \\
       \hline 
       Mutant-X \cite{alison} & Black-box &  Stylometric (lexical) &  & \cmark & TuringBench\\
       \hline 
       Avengers \cite{alison} & Black-box &  Stylometric (lexical) &  & \cmark & TuringBench  \\
       \hline
    \end{tabular}
    }
    \caption{Authorship Obfuscation techniques for Neural Texts. 
    \textit{With cited papers that implemented them}.}
    \label{tab:taxao}
    % \vspace{-10pt}
\end{table*}

\begin{figure}
    \centering
    \includegraphics[width=0.75 \linewidth]{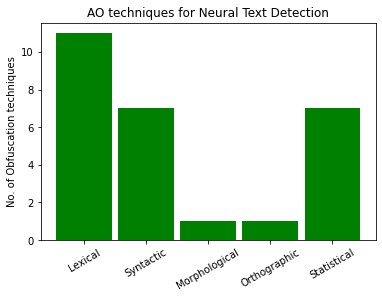}
\caption{Number of AO techniques for NTD per category in the Taxonomy}    \label{fig:ao_bar}
\end{figure}

\section{Authorship Obfuscation for \\Neural Texts} \label{authorshipobf}
In the task of NTD, AA models are evaluated 
under adversarial settings to assess their robustness. 
Due to the security risk, NTGs pose, it is important that AA models are robust to 
adversarial perturbations. 
The problem of administering adversarial perturbations to texts to cause an accurate AA model 
to assign inaccurate authorship is called \textbf{Authorship Obfuscation} (AO). 
This is because AO is the 
process of masking an author's writing style/signature 
to conceal the identity, usually for privacy reasons
\cite{mahmood2019girl}. 
It is a strict case of Adversarial attacks \cite{zhang2020adversarial}, as the 
goal is to obfuscate writing style and preserve semantics, such 
that both human and automatic detection is evaded. 
See the differences between the two
in Table \ref{tab:ao}. 
AO is a well-studied problem in the traditional AA community \cite{karadzhov2017case, mahmood2019girl, mahmood2020girl, zhai2022girl,mcdonald2012use},
and has been extended to the niche AA community for NTD. 
This AO for neural texts problem is formulated as: 

\begin{myBox}[]{Definition}{AO}
\textsc{\textbf{Definition of AO for Neural Texts}}. \textit{Given an AA model $F(x)$ that accurately assigns authorship of text $t$ 
to $k$ which can be either a human or an NTG author,
the AO model $O(x)$ slightly modifies $t$ to $t^*$ (i.e., $t^*{\leftarrow}O(t)$) such that the authorship is disguised (i.e., $F(t^*){\neq}k$) and
the difference between $t^*$ and $t$ is negligible.}
%($d(t^*, t)\approx0$), where 
%$d$ is a metric function that measures semantic similarity.}
\end{myBox}

% \thai{I would remove the cosine similarity, or make it very clear like cosine similarity of semantic vectors. I can re-write this definition later.}  
% AO has only been lightly studied with just a few works
% \cite{karadzhov2017case, mahmood2019girl, mahmood2020girl, zhai2022girl,mcdonald2012use} 
% while \textit{Adversarial Attacks} 
% has been heavily studied with a long list of works\footnote{https://github.com/thunlp/TAADpapers}. 
% We discuss the landscape of AO techniques applied to the AA models for 
% NTD. 
Thus, we survey all AO techniques employed to obfuscate neural texts
in different categories, as illustrated in Figure \ref{fig:aotax}:
 % in 3 categories - 
 \textit{Stylometric Obfuscation},
and \textit{Statistical Obfuscation}. 
% These categories represent the features used to obfuscate neural texts. 
% However, there are other categories these AO techniques can be further divided 
% into - \textit{attack method} and \textit{attack scenario}. 
% There are two kinds of attack methods - direct and non-direct attack. 
% The direct attack method needs access to either the classifier that will be 
% used for the AA task or a substitute classifier,
% while the non-direct attack does not need it. 
% There are several constraints within AO category that is used to 
% obfuscate texts. These include POS, perplexity, n-grams, etc. 
% Finally, we only use the feature-based categories for the taxonomy of AO 
% techniques because it is a more comprehensive description of the AO techniques. 
% These categories are able to capture the nuances of each AO algorithm.
% See Figure \ref{fig:aotax} for a flowchart of the taxonomy
% of AO techniques used to obfuscate neural texts. 
% See Figure \ref{fig:aotax} for a flowchart of a taxonomy for AO and 
% Table \ref{tab:taxao} for a list of AO techniques used to obfuscate neural texts. 

\subsection{Stylometric Obfuscation}
In order to build a robust stylometric classifier, as is observed in 
Section \ref{stylo}, an ensemble of features that capture several 
linguistic structures such as - Lexical, Syntax, etc. are required. 
However, to obfuscate an author's writing style, only one 
of the linguistic structures may be perturbed. Therefore, 
all the stylometric AO techniques only target a specific 
linguistic structure, unlike AA classifiers. 
Based on the stylometric obfuscation 
techniques used to obfuscate neural texts, we further divide this category into 4 categories - \textit{Lexical, Syntactic, Morphological,} 
and \textit{Orthographic} Obfuscation.

\subsubsection{Lexical Obfuscation}
% \thai{Maybe briefly define lexical--e.g., dealing with or related to the use of words/words choice, etc.}
Lexical relates to the word choice of a piece of text. Thus lexical obfuscation algorithms 
aim to mask authors' writing styles by replacing certain keywords with their synonyms 
while preserving semantics. Below, we discuss different techniques used to achieve lexical obfuscation for neural texts. 
Misspellings attacks may be considered a trivial AO technique, 
however, it is effective in obfuscation.
The misspellings attack uses a list of commonly misspelled words\footnote{\url{https://en.wikipedia.org/wiki/Wikipedia:Lists\_of\_common\_misspellings/For\_machines}}
to determine which words to replace with their misspelled version.
This AO technique is successful in obfuscating texts generated by GPT-2 and GROVER, 
and thus, evades detection of the following AA models - 
\textit{GLTR}, \textit{GROVER}, and \textit{GPT-2} detector \cite{wolff2020attacking}. 
% It was found to be as successful at the homolgyph attack.
\textit{GROVER detector} is further evaluated with this AO technique
by obfuscating texts generated by GROVER. 
% \thai{let's change GROVER texts -> texts generated by GROVER. Similarly with GPT-2, etc.}. 
With only less than 10\% of the texts perturbed, this attack is 
94\% successful \cite{gagiano2021robustness}. 
However, this attack can be maneuvered by spell check algorithms, 
making the obfuscation technique, not robust \cite{wolff2020attacking}. 
In addition to misspelling, a whitespace attack
(``will face" $\to$ ``willface") is used to evaluate the robustness of
\textit{GROVER detector}. With only less than 4\% 
of the texts perturbed, the attack is 85\% successful \cite{gagiano2021robustness}. 

Interesting artifacts/characteristics of neural texts still remain somewhat elusive.
Therefore, perturbing these neural texts could reveal characteristics that have 
evaded the AA \& AO community. 
Hence, using linguistic and statistical perturbations of words in the text, \cite{puunraveling} extract important characteristics of neural text.
For the linguistic-based perturbations, a lexical obfuscation technique is implemented - 
\textit{Deduplicate tokens} which keeps the first 
occurrence of a token/word as is and replaces other 
occurrences with \textit{[MASK]} token. This AO technique surprisingly improves 
the AA performance, suggesting that reducing the number of token occurrences
may remove trivial features, causing the AA classifiers to focus on the 
more important features \cite{puunraveling}

Next, texts generated by GROVER are obfuscated with the following techniques: 
(1) \textit{varying sentiment}: changing the sentiment of words by replacing the word with another word of a different sentiment (positive $\to$ negative); and
(2) \textit{entity replacement}: replace entity with a useless entity \cite{bhat2020effectively}.
Results suggest that both  \textit{GROVER} and \textit{GPT-2} \textit{detectors} are 
vulnerable to these 
lexical-based perturbations. 

Mutant-X \cite{mahmood2019girl}
and Avengers \cite{haroon2021avengers} are used 
as baseline AO techniques to obfuscate the TuringBench dataset \cite{alison}. 
Mutant-X uses a genetic 
algorithm to search for suitable word replacement such that the semantics are 
preserved and the internal/substitute AA model misclassifies \cite{mahmood2019girl}. This process is notorious for its expensive computational complexity \cite{tabatabaei2015survey}.
An internal model is used because, in the real world, the original AA model may not be known. Moreover,
Mutant-X generates the obfuscated text and tests if it has evaded 
the AA model. If evasion is not successful, the process is repeated and tested 
for the defined max number of iterations \cite{mahmood2019girl}. 
These factors significantly increase the runtime of Mutant-X. 
Furthermore, the success of Mutant-X is dependent on how strong the internal 
AA model, which undermines the generalizability of Mutant-X. Hence, Avengers
\cite{haroon2021avengers}, an improved version of Mutant-X is proposed.
Avengers is an ensemble version of Mutant-X. Unlike, Mutant-X, the 
internal AA model is an ensemble AA model, such that each classifier
out of $N$ classifiers focuses on different linguistic structures - 
syntax, semantics, etc. 

DeepWordBug \cite{gao2018black}, a realistic 
character-level black-box attack 
is used to evaluate the robustness of 
3 types of model - Statistical classification model \cite{nguyen2017identifying}, 
RoBERTa \cite{liu2019roberta}, and an Ensemble model 
(Statistical model + RoBERTa) \cite{crothers2022adversarial}. 
It perturbs characters such that misclassification is 
maximized, while the Levenshtein edit distance is minimized 
\cite{gao2018black, crothers2022adversarial}.  
These models were trained with GPT-2 medium webtext and 
tested 3 separate test datasets -  human vs. neural webtext 
from GPT-2 medium, GPT2 XL, and GPT-3 \cite{crothers2022adversarial}. 
While deep learning-based classifiers achieve a higher 
performance in unperturbed/clean texts, statistical classifiers 
were found to be more robust to obfuscation \cite{crothers2022adversarial}. 
Thus, the Ensemble model merges the advantages of high performance and adversarial 
robustness of the 2 models.
DeepWordBug did not reasonably degrade the Ensemble
model's performance, suggesting that DeepWordBug is 
not robust for this task. 
However, when DeepWordBug is used to evaluate the robustness of 
\textit{GROVER detector} 
(Mega model) and \textit{OpenAI  detector} 
(base and large models) by perturbing the GPT-2 generated Yahoo answers \& Yelp Polarity vs. 
Human datasets, it is successful
\cite{stiff2022detecting}. 
In fact, DeepWordBug is found to be a 
very successful AO technique, reducing the accuracy of the Yahoo and Yelp datasets 
from 67.9\% to 0.4\% and  87.4\% to 6.9\%, respectively
\cite{stiff2022detecting}. 
This suggests that the \textit{OpenAI} and 
\textit{GROVER detectors} are not robust to this kind of AO technique when 
evaluated on GPT-2 generated Yahoo answers \& Yelp Polarity. 

In addition, TextFooler \cite{jin2020bert},
a realistic word-level
black-box attack is used to evaluate the robustness of the
Statistical model, 
RoBERTa, and Ensemble model 
(Statistical model + RoBERTa) \cite{crothers2022adversarial}. 
TextFooler 
replaces words with synonyms based on cosine similarity within
the embedding space \cite{jin2020bert, crothers2022adversarial}. 
Based on the results, 
TextFooler is a robust AO technique, 
especially for Transformer-based models. Furthermore, 
as a substitute for human judgment, \textit{MAUVE} is used to 
measure the human judgment of obfuscated texts.
\cite{crothers2022adversarial} finds that adversarial perturbation 
reduces \textit{MAUVE} score which means that text quality is 
degraded and therefore neural texts are likely 
to be detected accurately by humans. 

To further evaluate the robustness of the AA models - 
\textit{GLTR} (\textit{GLTR-BERT} \& \textit{GLTR-GPT2}),  
\textit{GROVER detector}, \textit{BERT-Defense}, \textit{FAST}, 
and \textit{RoBERTa-Defense}, 
texts generated by GPT-2 and GROVER are obfuscated with 
TextFooler, Random Perturbations \cite{pudeepfake}, and DFTFooler \cite{pudeepfake} AO techniques. 
Random Perturbations is an 
attack method that replaces random words with synonyms while preserving the semantics. 
DFTFooler is similar to TextFooler 
but only needs a publicly available pre-trained LM to generate obfuscated texts. 
This makes DFTFooler not as computationally costly as TextFooler \cite{pudeepfake}. 
Also, DFTFooler perturbations are transferable \cite{pudeepfake}. 
To find a valid word substitution using DFTFooler, there are 4 steps:
(1) synonym extraction; 
(2) POS checking;
(3) Semantic Similarity checking; and 
(4) Choose a synonym with low confidence as measured by a LM. 
BERT and GPT-2 XL are used as the LM for DFTFooler. 
Results suggest that TextFooler is a stronger AO technique than 
DFTFooler, but performs comparably, achieving 23-91\% Evasion Rate \cite{pudeepfake}. 
Evasion rate is defined as the fraction of
perturbed neural text that evades detection by an AA model.
A high evasion rate indicates a high attack success.
% DFTFooler is not as computationally heavy as TextFooler \cite{pudeepfake}. 
Furthermore, using a bi-directional LM (BERT) as the backend for DFTFooler,
and increasing the number of words perturbed,
increases the evasion rate of DFTFooler. 
Based on the results, 
\textit{FAST} is the most adversarially robust AA model. This may be due to 
the hybrid nature of the model as it combines 
the benefits of stylometric, statistical, and deep learning-based techniques as discussed in section \ref{AA}. Another reason for \textit{FAST}'s superior performance is its use of semantic features based on entities \cite{pudeepfake}.
% \cite{pudeepfake}'s claims that \textbf{FAST}'s use of semantic features based on entities could be the reason for its robustness. 

\subsubsection{Syntactic Obfuscation}
% \thai{Briefly explain what is syntactic--e.g., structure, ordered placement of words, etc.}. 
Syntax relates to the order of words in a piece of text. 
Thus, syntactic obfuscation techniques change the original arrangement of words in a document, 
in order to obfuscate the author's writing style. 
Below, we discuss such techniques used on neural texts.
Characteristics of neural texts are extracted 
by perturbing the syntactic structure of the texts with the following syntactic perturbation techniques: 
(1) \textit{Shuffle tokens} which randomly shuffles the word order 
of the texts;
(2) \textit{Retain only (non)-stopwords} which removes all words, 
except for stopwords \cite{puunraveling}.  
The accuracy of the AA model only dropped marginally. 
This suggests that these AO techniques are not robust. It also implies 
that these syntactic features are not important for NTD. 

The robustness of \textit{GROVER detector} is further evaluated by 
syntactic obfuscation techniques on texts generated by GROVER. 
These techniques are: 
(1) \textit{source-target exchange}: interchanging the source and target;
(2) \textit{alter numerical facts}: distort numerical facts;  
(3) \textit{syntactic perturbations}: changing the word form by adding/removing punctuation (``There is" $\to$ ``There's"); and 
(4) \textit{article shuffling}: replacing 
$N$\% of a real (human-written) article's sentences with $N$ 
sentences of a fake (neural) article \cite{bhat2020effectively}.
All AO techniques were successful, except \textit{article shuffling}. 
Also, stylometric classifiers were found to be more robust, 
except when perturbed under stricter constraints, 
such as perturbing a large percentage of texts \cite{bhat2020effectively}. 
% \textit{Article
% shuffling} because  \textit{GROVER detector} is insensitive to \textit{article
% shuffling}. 

% Finally, \cite{alison}, proposes an AO technique - \textit{ALISON}. 
\textit{ALISON} \cite{alison} is another syntactic AO technique. 
It reduces inference time by 100-200x when compared to SOTA AO techniques 
such as Mutant-X \cite{mahmood2019girl}, and 
Avengers \cite{haroon2021avengers}. 
% Unlike most of the SOTA AO techniques, 
% ALISON does not require a substitute model to generate obfuscated texts. 
It has an internal classifier trained on a set of linguistic AA features, which allow
ALISON to generate suitable phrase replacements that preserve semantics. 
ALISON is used to evaluate the robustness of 3 Transformer-based models - BERT, DistilBERT, and RoBERTa   
by obfuscating 2 datasets - TuringBench and Blog Authorship Corpus \cite{alison}.  
It is able to obfuscate the datasets well, causing the AA models to underperform on obfuscated texts. Furthermore, ALISON is able to preserve the semantics 
of the original text much better than their baseline AO techniques (i.e., Mutant-X and Avengers). 
% The similarity between original and obfuscated texts is measured with 
% METEOR and USE (Universal Sentence Encoder) cosine similarity metrics. % thai I dont think this sentence is needed here.

\subsubsection{Morphological Obfuscation}
Morphology is the study of word forms. Thus, morphological obfuscation techniques
change the configuration of a word (e.g. ``won't'' $\to$ ``will not'').
Upper/Lower Flip (``Leaving" $\to$ ``LeavIng") is a morphological AO technique
that may be considered trivial. However, it is successful in 
obfuscating texts generated by GROVER which significantly reduces the performance of 
\textit{GROVER detector} 
\cite{gagiano2021robustness}.
With only about 2\% of the texts perturbed, it
achieved a 96\% success rate in evading \textit{GROVER detector}'s
detection \cite{gagiano2021robustness}. 

\subsubsection{Orthographic Obfuscation}
% \thai{what is orthographic?}. 
Orthography is the spelling convention of a language. 
Thus, orthographic obfuscation techniques aim to change the original spelling convention 
used in a piece of text to mask an author's writing style. 
Below, we discuss such techniques. 
Homoglyph attack is an orthographic perturbation technique 
that changes the unicode of texts. It changes English characters to cyrillic characters. 
This attack is almost 
imperceptible to the human eye and therefore, preserves semantics. 
The robustness of 
\textit{GPT-2 detector}, \textit{GROVER detector}, and 
\textit{GLTR} is evaluated by obfuscating texts generated by GPT-2 with 
the homoglyph attack.
% Homoglyph attacks change the 
% unicode of selected characters in a word. 
% This task involves changing English characters to Cyrillic characters. 
% The goal of this attack is to change 
% the distribution of generated texts, such that neural text detectors
% misclassify. 
\textit{GPT-2 detector}'s performance dropped from 97.44\%  to 
0.26\% Recall.  The perturbed texts caused \textit{GROVER detector} 
to grossly misclassify neural texts as human-written texts 
and \textit{GLTR} to shift 
the distribution (i.e., color scheme) of the perturbed texts \cite{wolff2020attacking}.
\textit{GROVER detector} is further evaluated on obfuscated texts generated by GROVER
\cite{gagiano2021robustness}. Homoglyph attack achieves a 97\% success rate in perturbing \textit{GROVER detector} \cite{gagiano2021robustness}.
However, this AO technique can easily be rendered ineffective 
by using spell check algorithms \cite{wolff2020attacking}.  
% Using 
% the Upper/Lower Flip (``Leaving" $\to$ ``LeavIng") attack with 
% only about 2\% of the texts were perturbed, and the algorithm was 96\% 
% successful. 
% \subsection{Deep Learning Obfuscation}

\subsection{Statistical Obfuscation}
In order to extract statistical characteristics from neural texts, 
the following statistical AO techniques are implemented:  
(1) \textit{Replace text with likelihood ranks}; 
(2) \textit{Replace text with specific linguistic features, 
such as Part of Speech, Dependency Trees,
Constituent Trees} and \textit{Named Entities}; and
(3) \textit{Retain tokens in high/low frequency regions} which 
defines a frequency gap score to calculate and extract the high and 
low-frequency words in the text \cite{puunraveling}. 
The following 3 datasets are perturbed - human-machine pairs, 
Writing Prompt dataset \cite{fan2018hierarchical},
and CnDARIO (Chinese Novel Dataset crAwled
fRom mIxed online sOurces) generated with GPT-2 fine-tuned with Chinese Literature. 
These datasets are in 3 different domains, respectively - 
Online Forum, News, and Literature. 
Results suggest that 
the \textit{high/low frequency region} perturbations is the most effective obfuscation technique \cite{puunraveling}. 
This further suggests that the \textit{high/low-frequency region} feature could 
be an effective feature for distinguishing 
neural texts from human texts. 

\textit{ruBERT} for NTD is evaluated on  
2 AO techniques -
(1) \textit{calculating the 
class probabilities of each label and only selecting the texts 
with the highest human class probability}; 
(2) \textit{adds the detector's 
log-probability to the beam search decoding strategy 
during generation} so that 
only more human-like texts are generated
\cite{automik22}. These attacks achieve
46\% and 56\% success rates, respectively. 
\textit{RoBERTa-Defense} is evaluated 
by changing the sampling distribution of the neural texts in the test set. 
This obfuscation technique involves: 
(1) \textit{varying the text decoding strategy (and its parameters)}; and 
(2) \textit{varying the number of priming tokens} \cite{pudeepfake}. 
The quality of the obfuscated neural texts is measured by GRUEN \cite{zhu2020gruen}, 
a metric used to measure the linguistic quality of AI-generated texts 
(neural texts). GRUEN has a high correlation with human judgments.
The score ranges from $0-1$, and a higher value indicates high linguistic quality. 
Linguistic quality is based on grammaticality, non-redundancy, discourse
focus, structure, and coherence \cite{pudeepfake}. 
Using GRUEN, a \textit{successful attack} is defined as an 
attack that degrades the performance of the AA models, 
with little to no linguistic quality (GRUEN score)
degradation \cite{pudeepfake}.  
The results suggest that changing the decoding strategy is an effective AO technique. 
Even \textit{GLTR}, a statistical AA model is fooled by this AO technique \cite{pudeepfake}. 
% Also, the AA models trained on conditionally generated texts and tested on 
% unconditionally generated text, underperform. This is evident 
% in \textbf{BERT-Defense}'s performance which shows over 47\% degradation 
% in performance \cite{pudeepfake}. 

% character-level 
% Deepword bug

% word-level 
% varying sentiment 
% source-target exchange 
% entity replacement
% syntactic perturbation
% Textfooler
% deduplicate tokens, shuffle tokens, remove stopwords, 
% Retain tokens in high/low frequency regions

% sentence-level 
% article shuffling
% alter numerical facts

% Syntactic
%  (i) breaking longer sentences, 
% (ii) removing definite articles if they appear among the most repeated words in an article, 
% (iii) using semantic-preserving rules (for example converting that’s →that is) (Ribeiro et al., 2018), 
% (iv) reformatting paragraphs of machine-generated text.

\section{Evaluation of AA/AO Methods} \label{eval}

%In this section, we summarize how literature evaluates the methods of AA/AO. Roughly speaking, in the machine-based evaluation, one uses  

\subsection{Machine-based Evaluation}

\subsubsection{Authorship Attribution}
To evaluate AA models, literature often used popular classification metrics such as  \textit{Precision}, \textit{Recall}, \textit{Accuracy}, and \textit{F1 score}. 
%The most popular metrics are the F1 score and Accuracy. 
For instance, \cite{wolff2020attacking} used the recall metric to evaluate the robustness of AA models toward AO techniques. 
Previous works evaluate the generalizability of AA models, not only on a standard single test set \cite{holtzman2019curious,pudeepfake,frohling2021feature,ippolito2020automatic,munir2021through} 
but also on several out-of-sample distributions \cite{stiff2022detecting}. For example, \cite{stiff2022detecting} evaluate \textit{GROVER detector} (Mega model) and 
\textit{OpenAI  detector} (base and large models)
on three variants of test sets, namely 
\textit{in-distribution, out-of-distribution} and 
\textit{in-the-wild} datasets. 
\textit{In-distribution} datasets are test sets sampled from the same distribution of the training set, while
\textit{Out-of-distribution} datasets are those sampled from a different distribution from the training set. 
This test dataset is created using PPLM \cite{dathathri2019plug} and GeDi \cite{krause2021gedi} to control the GROVER and GPT-2 generations \cite{stiff2022detecting}.
The \textit{in-the-wild} datasets are test sets generated from an NTG, different from the NTG used to generate a training set \cite{stiff2022detecting}.
To build this \textit{in-the-wild} dataset,  training sets contain texts generated from GPT-2 and GROVER pre-trained models, while test sets contain texts generated from GPT-3 and a fine-tuned GPT-2 model \cite{stiff2022detecting}.
%In addition, the Authorship Attribution-AA dataset is included in this  \textit{in-the-wild} test set \cite{stiff2022detecting}.
% \thai{I am not quite understand the last sentence. Can you check?}. 
% The goal here is to create a real-world scenario where AA models detect neural texts,
% authored by NTGs not seen in the train set. 
In general, AA models perform well on \textit{in-distribution} datasets, but suffer on \textit{out-of-distribution} datasets, and even more on \textit{in-the-wild} datasets.

%dongwon--unclear what you wanted to say below, so commented it out.

%All evaluated AA models perform relatively well on \textit{in-distribution} datasets. However, for the \textit{out-of-distribution} datasets, they perform much better at detecting PPLM than GeDi texts. This suggests that GeDi may generate more superior and harder to distinguish texts from humans. Finally, using the \textit{in-the-wild} datasets, the AA models perform very poorly. 

\subsubsection{Authorship Obfuscation}
To evaluate AO models, literature often uses the \textit{success rate} \cite{gagiano2021robustness,automik22}, a fraction of successfully
obfuscated (i.e., misclassified) texts which were accurately attributed prior to obfuscation.
Another measure for AO models is \textit{evasion rate} \cite{pudeepfake} which is the fraction of perturbed neural texts that evade the detection by an AA model.
% \lee{add a few citations of AO mdoels where both measures--success rate and evasion rate--were used}
%A high \textit{success} and \textit{evasion} rate indicates a  successful obfuscation. 
Next, due to the time and financial cost required to carry out a human-based evaluation,  \textit{MAUVE}, 
a metric that statistically emulates human judgments in terms of the linguistic quality (i.e., coherency) of neural  texts vs. human-written texts, has been used on the AO problem \cite{crothers2022adversarial}. 
That is, \textit{MAUVE} is used as a substitute 
for human evaluation of obfuscated text vs. non-obfuscated text \cite{crothers2022adversarial}. 
Furthermore, based on the strict definition of AO, it is sometimes
important that the obfuscated text preserves the semantics of the original text. 
Hence, literature has measured the degradation of 
semantics between original and obfuscated texts, namely 
METEOR~\cite{banerjee2005meteor}, 
Universal Sentence Encoder (USE)~\cite{cer2018universal} Cosine similarity, and 
GRUEN~\cite{zhu2020gruen}. 
These metrics all correlate with human judgments and a high score indicates an obfuscated text with well-preserved semantics.

\subsection{Human-based Evaluation} \label{human-based}
% \thai{similar to 5.1, I suggest we divide to AA and AO task. Or, you can have section 5.1 evaluation AA and section 5.2 evaluation AO.}
% \subsubsection{Authorship Attribution} 
% \thai{Section 5.2 only has 1 subsection and it is rather long. How about we splitting Section 5.2 into something like ``Human Evaluation without Training", ``Human Evaluation with Training". The last paragraph of this section then will fit into this structure.}
There is currently no known human-based evaluation of AO models. The closest one is the machine-based simulation by MAUVE \cite{crothers2022adversarial}. As such, in this section, we focus our review on the human-based evaluation of AA models, especially in the context of NTD.

%is 
%currently the closest metric to a human-based 
%study in the AO problem. 
%This means that all other human-based evaluations have 
%only been used in the AA context, specifically, 
%the \textit{Turing Test} special case where $k=2$ authors - 
%human vs. machine.
%Therefore, we categorize the human-based evaluations into 
%\textit{Human Evaluation without Training} and 
%\textit{Human Evaluation with Training}.

\subsubsection{Human Evaluation without Training}

A set of research works recruited human participants (often from crowdsourcing platforms such as Amazon Mechanical Turk (AMT)) and tested whether they can distinguish neural texts from human-written texts. A simple introduction to the tasks is given, but no special training is done for human participants in this line of research.
For instance, \cite{zellers2019defending} examined the quality of texts generated by GROVER vs. human-written texts  by humans and found that humans find GROVER-written news more believable than human-written news.
%Particularly, humans were asked to choose which news articles they found to be more believable between texts generated by GROVER and human-written news articles \cite{zellers2019defending}. 
%Surprisingly, results suggest that humans  find GROVER-written news more believable than human-written news.
\cite{donahue2020enabling} asked human participants to detect infilled texts filled
by neural texts (e.g., she drank [blank] for [blank]) and found that humans had difficulty in detecting the infilled texts filled by neural texts.
\cite{uchendu2021turingbench} introduced a benchmark for AA research, \emph{TuringBench}, evaluated the performance of humans in distinguishing 19 pairs of human vs NTG (e.g., human vs. GPT-1 or human vs. FAIR) using TuringBench, and found that humans on average scored 51-54\% of accuracies, only slightly better than random guessing.
%They conducted two human-based studies - (1) given one article, the participants were asked to vote if the article is machine- or human-generated; (2) given two articles, one is human-written and the other machine-generated, the participants were asked to choose the machine-generated article. Humans underperform in the task, achieving a 54\% and 51\% accuracy, respectively, which is comparable to random guess - 50\%. 
%Unknown to the participants, all of the articles for the first study were actually machine-generated \cite{uchendu2021turingbench}. 
Unsurprisingly, \cite{brown2020language} also found that humans were unable to 
accurately detect GPT-3 texts from human-written texts.
% and find that humans find it difficult to detect GPT-3 texts \cite{brown2020language}. 
\cite{ippolito2020automatic} evaluated the quality of top-p and top-k decoding strategies, and found that
(1) AA models detect neural texts generated by top-k decoding better than humans, but (2) humans detect neural texts generated by top-p decoding better than AA models.
%for texts generated using top-k strategy, AA models found it easier to detect than humans, and (2) for texts generated using top-p strategy, are easier for humans but harder for AA models to detect \cite{ippolito2020automatic}. 
%The reason is that top-k strategy follows a distribution that is  easy for AA models but difficult for humans to capture, and vice-versa for top-p strategy \cite{ippolito2020automatic}. 
Lastly, \cite{adelani2020generating} found that both humans and AA models 
struggled to detect neural fake reviews.

\subsubsection{Human Evaluation with Training}

Another line of research attempted to first train human participants about NTD tasks and measure the performance improvements afterward.
For instance, when human participants were trained to use GLTR \cite{gehrmann2019gltr} in detecting neural texts, 
%to evaluate the interpretability of \textit{GLTR}  \cite{gehrmann2019gltr}, 
%human participants were trained to use \textit{GLTR} to assist them in 
%detecting neural texts. 
thanks to the color scheme of GLTR (see Figure \ref{fig:gltr}), 
human performance increased from  54\% to 72\% in accuracy.
To further evaluate neural texts, \cite{dugan2020roft} proposed
a framework to collect a large number of human annotations via a game, Roft\footnote{http://roft.io/},
on the quality of neural vs. human texts.
Human participants were told to detect the boundary at which an article
transitions from human-written to neurally-generated. 
Only 16\% of human participants were able to correctly identify the accurate boundaries \cite{dugan2020roft}.
% \lee{what's the conclusion of this work ROFT? using game, how much did it improve it? summarize it as 1 line here, not verbosely.}
%It was set up as a game\footnote{http://roft.io/} where a user chooses which category they want, and is asked to select if a sentence is human-written or machine-generated. 
%If human-written is selected,
%another sentence with the same question is given. However, if machine-generated is selected, a series of potential reasons why the selection could be machine-generated is
%presented, for example., repetition, boring to read, etc. 
\cite{clark2021all} studied three training strategies--instruction-based, example-based, and comparison-based, and found that example-based training is the most effective to improve human performance for solving NTD tasks (achieving the average accuracy of 55\%) across the domains of story, recipe, and news.
%To improve human performance in NTD, 
%three training frameworks: 
%\textit{Instruction-based, Example-based,} and \\ \textit{Comparison-based} training
%is used to train human participants \cite{clark2021all}. 
%Human participants were only able to slightly 
%improve the accuracy of NTD.  
%\textit{Example-based training} performs the best with an average 
%performance of 55\% accuracy across all three domains - Story, Recipe, 
%and News \cite{clark2021all}. 
%This further confirms that improving human performance in NTD
%is very difficult and still remains an open problem \cite{uchendu2021turingbench,gehrmann2019gltr,zellers2019defending}. 
Next, 
\cite{tan2020detecting} investigated how accurately humans can  classify real vs. generated articles with and without images, using different types of news datasets: real captions and articles, real captions and generated articles, generated captions and real articles and generated captions and articles.
%the different combinations of the neural news dataset - 
%\textit{(A)} \textit{Real Articles and Real Captions}; 
%\textit{(B)} \textit{Real Articles and Generated Captions};
%\textit{(C)} \textit{Generated Articles and Real Captions};
%\textit{(D)} \textit{Generated Articles and Generated Captions}
%are evaluated by humans. 
% Next, using the different types of datasets (\textbf{A-D}) 
% in the NeuralNews dataset (see Table \ref{tab:data}), 
By conducting an AMT-based study, untrained and trained human participants were able to achieve an average of 46\% and 68\%, respectively.
%Also, the human participants achieved an average accuracy of 59\% on detecting just the human vs. neural articles without images. 
Further investigation on the trustworthiness of the different article types
based on style, content, consistency, and overall trustworthiness
reveals that humans were 
skeptical about the overall trustworthiness of news articles across all four types
\cite{tan2020detecting}.
Finally, recently, \cite{dou2021scarecrow} proposed a framework for scrutinizing neural texts through 
crowdsourced data annotation in a scalable fashion, where neural texts were shown to have various error types: language-errors (i.e., lack of coherency and consistency in text), factual-errors, and reader-issues (i.e., text is too obscure or filled with too much jargon so that understanding is negatively impacted).

In conclusion, literature has found that humans alone cannot detect neural texts accurately, achieving detection accuracies only slightly better than random guessing. When humans are properly trained about the characteristics of neural texts, further, this detection accuracy tends to increase but only by small margins.
%without any training cannot detect neural texts accurately.
%We observe that even in different domains (not just news), humans still struggle 
%to distinguish between neural and human-authored texts. 
%Furthermore, recently proposed human training procedures, only improve 
%human performance marginally. 
%Therefore, 
%we need better training methods to train humans, as well as human-in-the-loop neural text detectors to improve human detection performances. Using human-in-the-loop 
%neural text detectors, such as \textit{GLTR}, linguistic cues that are 
%interpretable to humans can be highlighted within a text, making it visually easy 
%to distinguish neural texts from human texts. 

\section{Open Challenges} \label{open}
Although there have been several meaningful works on the current landscape of AA and AO models, the two research problems are still in their early development, especially the direction of NTD. In this section, as such, we discuss some of the remaining challenges.
% Thus, below are the current open problems in the field of NTD:

\subsection{Need for Comprehensive Benchmark}
Generally, existing literature tends to create or use particular data-sets in silos, making their findings limited and incomparable across the literature. 
As mentioned in \cite{munir2021through,frohling2021feature}, however, the study of NTD can be greatly improved with the availability of more comprehensive and generalizable datasets whose coverage varies across diverse: (1) domains (e.g., news, online forum, recipe, stories), (2) language models, (3) decoding strategies, or (4) length of texts. Further, not all AA/AO models share their codebase and experimental configurations, making the comparative analysis difficult. However, generating and maintaining a large number of neural texts across different settings cost significant resources and effort.
\cite{uchendu2021turingbench} attempted to propose a benchmark for AA research, \emph{TuringBench}, but it does not satisfy the needs fully. Therefore, it is greatly needed to develop a comprehensive benchmark with diverse datasets of AA/AO problems, along with the codebase of known methods in a unified environment, so that objective comparison can be carefully performed to  understand the pros and cons of existing solutions and brainstorm new ideas for improvement.

\subsection{Call for Complex AA/AO Variations}  

%\vspace{5pt}
%\noindent\textbf{Call to study more complex variations of Authorship Attribution of neural texts}: 
With the introduction of ``machine'' in writing high-quality texts, the set-up of ``authors'' in future scenarios can be more complex. For instance, one could generate a more realistic text using multiple language models in sequence (e.g., each language model improves upon the text generated by another language model in a previous step) or in parallel (e.g., each language model generates only parts of a long text). Symmetrically, it is also plausible to use multiple AO solutions in sequence or parallel to improve the overall performance of obfuscation. Yet another possible scenario is to think of ``human-in-the-loop'' attribution or obfuscation. For instance, would a team (of humans, of machines, or of humans and machines) outperform an individual (of human or machine) in solving AA or AO task?
To our best knowledge, there is no study of AA/AO for such complex scenarios. 
% \lee{is this true?}

%    Malicious users of NTGs can use clever ways to generate misinformation in order
%    to evade detection by the AA models. Some of these clever AA variations are: 
%    \textit{using $k$ NTGs to generate different parts of a single 
%    document, can the SOTA AA models accurately assign authorship to the different parts of the document?} 
%    Say $k=3$ with NTGs - GPT-3, GROVER, and Human. 
%    Other variations can include $K$ NTGs vs. $K$ humans - where $k>1$. We can start by using different news outlets as different human labels (CNN, Washington Post, etc.). 
%    These AA questions 
%    are currently an open problem in the AA community, as detecting 
%    multiple authors in a document and then naming these authors 
%    is a 2-fold non-trivial problem that has no sufficient solution 
%    yet. Additionally, including deepfakes in these datasets will make an even more 
%    realistic scenario and exacerbate the difficulty of the problem. 
%    Nevertheless, this is a worthwhile task to investigate because currently, 
%    the field of NTD is playing catchup to the NTG field, thus,
%    making AA models obsolete with newer NTGs. Therefore, if we study harder
%    variations of the AA problem, we can potentially stay ahead of the NTG field
%    and stand the test of time. 

\subsection{Need for Interpretable AA/AO Models}  

%\vspace{5pt}
%\noindent\textbf{Need for more interpretable \& intuitive AA models}:
Currently, there are only a few interpretable AA models (e.g., GLTR)
and AO techniques (e.g., Homoglyph) for neural texts, as summarized in  Tables \ref{tab:detectors} and \ref{tab:ao},
respectively. 
That is, when an AA model detects a text as machine-generated or human-written, or when an AO model modifies parts of a text to hide authorship, it often cannot explain ``why?'' Ideally, however, such models should be able to provide an easy-to-understand and intuitive explanation, especially to users with no linguistic expertise, as to why a given text is attributable to a particular NTG or why a particular phrase of a text is critical to reveal an author's identity.
In addition, more research is needed to develop an intuitive human interface or visualization toward explainable AA/AO models.

%However, among these models, only one 
%    is intuitive and can be used by humans to improve their detection of neural texts
%    - \textit{GLTR}. As NTGs evolve, we observe that \textit{GLTR}'s 
%    initial impressive performance degrades significantly \cite{uchendu2021turingbench,wolff2020attacking}. This suggests that we need more
%    human-in-the-loop
%    interpretable \& intuitive AA models that can improve human performance in 
%    NTD. As discussed above, the use of graphical features 
%    such as TDA has interpretable capabilities. Thus, all that is needed, is to 
%    make TDA explanations intuitive to humans. This can be done in a straightforward way by adopting the color scheme of \textit{GLTR}. But for a more complex solution, 
%    an end-to-end model needs to be built first that takes in a text and outputs TDA weights and vice-versa. Second, an AA model built with TDA features to 
%    attribute neural texts is needed. 
%    Third, using the end-to-end and the TDA-based AA model as well as Explainable algorithms such as 
%    LIME \cite{ribeiro2016should}, 
%    the process can be reversed such that the focus of the TDA classifier is highlighted 
%    in the text. Next, to further gain insights into the characteristics of neural vs. human texts, \cite{puunraveling} perturbations techniques could be adopted to assess
%    what linguistic cues humans focus on (even when text is perturbed). 

\subsection{Need for Improved Human Training}

In parallel to improving the performance of AA/AO solutions, it is equally important to raise the awareness of AA/AO problems in the presence of neural texts, and to be able to train human users  to detect neural texts better (e.g., identify phishing or misinformation message that includes neural texts as parts) or use AO solutions to hide one's authorship (e.g., an activist posting his/her message on social media without revealing true identity).
As we illustrate in Section \ref{human-based}, however, humans are not good at detecting neural texts, and there are not many AA/AO solutions suitable for novice users to benefit from in solving AA/AO tasks. Worst, still, is that even a few AA models such as \textit{GLTR} that were shown to be able to help human users to detect neural texts better have become less effective with the advancement of NTGs. 
Therefore, great needs exist to have a better way to train human users in solving AA/AO tasks.

\subsection{Call for Robust AA/AO Solutions}
%\subsection{AO Open Problems}
% \thai{I think we can keep (2), }
% \begin{enumerate}
    %  \item \textbf{AO of $k>2$ NTGs vs. 
    % Human-written texts}: In Section \ref{authorshipobf}, we see that 
    % the AO problem has mostly been applied 
    % to the \textit{Turing Test} special case. \cite{alison} is the 
    % only work that has applied AO to $k>2$ authors. 
    % This leaves a relatively unexplored area of 
    % application for the $k>2$ authors. 
    % We may find that 
    % while GPT-3 is one of the hardest NTG to detect 
    % \cite{uchendu2021turingbench,clark2021all,brown2020language}, it is just as 
    % easy as GPT-1 to obfuscate. Another interesting finding could 
    % be that when GPT-3 texts are obfuscated, it is mostly 
    % misclassified as GROVER base or XLM (the worse performing NTG), according to  
    % \cite{uchendu2020authorship,uchendu2021turingbench}.

    % \item \textbf{Adversarially Robust Neural Text Detectors}:
    % In section \ref{authorshipobf}, we surveyed all the articles that tested the adversarial robustness of all the 4 categories of neural text detectors. We find that based on results, only stylometric and hybrid (i.e. ensemble of stylometric \& Transformer-based classifier) classifiers \cite{crothers2022adversarial,puunraveling}. However, stylometic classifiers underperform compared to 

%\noindent\textbf{Call for robust AO techniques}:
    In section \ref{authorshipobf}, we surveyed all literature that evaluated the robustness of AA/AO models, and found that most existing AO techniques do not preserve the original semantics of text well and  thus cannot easily evade the attribution of AA solutions, especially human detection. Similarly, as we adopt more sophisticated hybrid approaches for AA tasks, successful AO attacks to hide authorship will become more challenging. Part of the reason for these vulnerabilities in existing AA/AO solutions is that the bulk of existing literature has studied either AA or AO problem in separation, thus greatly limiting their robustness against the other problem.
Therefore, to stay relevant and synonymous with a real-life scenario, both AA and AO solutions need to learn from each other, and co-train/co-evolve, as in a min-max optimization game.

\section{Applications} \label{app}
% \begin{enumerate}

\textbf{Deepfake Detection}:
Successful solutions for AA/AO tasks can be useful in many applications. For instance, 
recently, the generation of realistic AI-made images\footnote{https://thispersondoesnotexist.com/} and videos, so-called ``deepfakes'', have flooded the Web.
While most of these deepfakes are made for humor, some are malicious in generating misinformation, spreading political propaganda, or attacking individuals \cite{mirsky2021creation}. 
In literature, in particular,  \cite{tan2020detecting} studies the realistic scenario where real images would be paired with neural texts to increase the authenticity of a news article as well as evade detection. In such a setting, successful AA solutions can point out the non-human nature of neural texts to users or can be used to extract features of neural texts for downstream deepfake detection models.
\vspace{0.1in} \\
\textbf{Chatbot Detection}:
Another application is for AA solutions to detect suspicious messages (e.g., phishing or chatbot messages) that may have been (partially) generated by NTG. Similar to neural texts of news format, shorter or informal chatbot messages are also hard to discriminate when generated by machines \cite{shao2019reverse,uchendu2019characterizing}. 
An example of the state-of-the-art chatbot is ChatGPT \cite{chatgpt} which has been used 
to generate medical writings \cite{gao2022comparing,biswas2023chatgpt}, 
finance writings \cite{dowling2023chatgpt}, etc. These applications of ChatGPT
have also increased the likelihood of cheating in academic writing \cite{cotton2023chatting}. 
Thus, AA models for neural text detection will be beneficial in distinguishing chatGPT-generated texts from human-written texts. 
\vspace{0.1in} \\
\textbf{Anonymity Preservation}:
On the other hand, successful AO solutions can be used to help individuals who have needs to share their writings without jeopardizing their secret identity. For instance, an NGO activist or whistleblower may submit her op-ed to news media after making sure that no popular AA solutions can attribute the writing to her.

\section{Conclusion} \label{conclusion}

With the rise of neural texts that were generated by large-scale language models, 
we are currently in an arms race between generation and detection of \textit{neural texts}. In this work, we comprehensively survey two important problems of neural texts: \textbf{Authorship Attribution} (AA) and \textbf{Authorship Obfuscation} (AO). 
%In the AA problem, researchers mostly study the special case, where there are only 2 authors (human \& machine).
%However, there are still a number of AA models that solve the NTD problem for $k>2$ authors. 
% This special case is called - \textit{Turing Test}. 
We first categorize existing AA solutions into four types of
%Based on the proposed solutions to the AA problem, 
%we categorize these models into 4 categories - 
stylometric, deep learning-based, statistical, and hybrid attribution. Similarly, we categorize existing AO solutions into two types of stylometric and statistical obfuscation, and elaborate pros and cons of representative methods therein.
%Next, the AO techniques used to obfuscate neural texts are categorized into - 
%\textit{Stylometric}, and \textit{Statistical} Obfuscations. 
%
%dongwon--no obligation to survey arxiv paper as they are not peer-reviewed yet, and your claim below is rather weak, so having the statements below would weaken our submission. So, let's not mention them.
%Finally, as the field of NTD is still in 
%its infancy, there are 3 similar surveys \cite{jawahar2020automatic,crothers2022machine,guerrero2022synthetic} that discuss and provide new directions for the field. However, while they provide 
%insights, 
%we differ by discussing the problem through \textit{A Data Mining Perspective}, based on the landscape of AA and AO techniques.
In addition, we discuss different evaluation methods for AA and AO problems in the context of neural texts, and finally, share a few important challenges that we feel lacking currently.
By and large, we believe that the data mining community is well-positioned to be able to contribute to significant improvement in both AA and AO research. Despite their close implications in security and privacy, with respect to the underlying methods used, their problem formulation as supervised or unsupervised learning, and their focus on the accuracy and running time as major metrics. 
%In this survey, we identified several open problems and  potential applications to real life. 
%See Figure \ref{fig:open} for a Venn diagram of the open problems in  the AA \& AO problems and the problems that intersect these 2 topics. 
%We also discuss the data generation process briefly and the evaluation metrics used to evaluate the AA and AO techniques. Based on the literature, we categorize the evaluation metrics into \textit{machine-based} and \textit{human-based} evaluation techniques for AA and AO techniques. The \textit{human-based} evaluation is further categorized into - \textit{human evaluation without training} 
%and \textit{human evaluation with training}.
Lastly, to mitigate the challenges of accurate detection of neural text, \cite{kirchenbauer2023watermark} proposes watermarking 
these text-generative language models. This entails embedding 
humanly imperceptible signals into the language models such that 
they generate semantically relevant texts, unnoticeable to humans but 
noticeable to detectors. These watermarking \cite{kirchenbauer2023watermark,abdelnabi2021adversarial} techniques 
attempt to solve the security risks that these language models pose. 
However, as these watermarking techniques have not yet been widely adopted, we still have to rely on AA and AO solutions for neural text detection. Also, as such watermarking techniques are a recent/new development, 
their robustness to strong AO techniques has not yet been evaluated.

\section{Acknowledgment}
This work was in part supported by NSF awards 
\#1820609, \#2114824, and \#2131144.

%\newpage
% \bibliographystyle{ACM-Reference-Format}
\bibliographystyle{abbrv}
\bibliography{sample-base}

\begin{thebibliography}{100}

\bibitem{abbasi2008writeprints}
A.~Abbasi and H.~Chen.
\newblock Writeprints: A stylometric approach to identity-level identification
  and similarity detection in cyberspace.
\newblock {\em ACM Transactions on Information Systems (TOIS)}, 26(2):1--29,
  2008.

\bibitem{abdelnabi2021adversarial}
S.~Abdelnabi and M.~Fritz.
\newblock Adversarial watermarking transformer: Towards tracing text provenance
  with data hiding.
\newblock In {\em 2021 IEEE Symposium on Security and Privacy (SP)}, pages
  121--140. IEEE, 2021.

\bibitem{adelani2020generating}
D.~I. Adelani, H.~Mai, F.~Fang, H.~H. Nguyen, J.~Yamagishi, and I.~Echizen.
\newblock Generating sentiment-preserving fake online reviews using neural
  language models and their human-and machine-based detection.
\newblock In {\em International Conference on Advanced Information Networking
  and Applications}, pages 1341--1354. Springer, 2020.

\bibitem{constrabert}
B.~Ai, Y.~Wang, Y.~Tan, and T.~Samson.
\newblock Whodunit? learning to contrast for authorship attribution.
\newblock {\em Proceedings of the 2nd Conference of the Asia-Pacific Chapter of
  the Association for Computational Linguistics and the 12th International
  Joint Conference on Natural Language Processing}, 2022.

\bibitem{bakhtin2019real}
A.~Bakhtin, S.~Gross, M.~Ott, Y.~Deng, M.~Ranzato, and A.~Szlam.
\newblock Real or fake? learning to discriminate machine from human generated
  text.
\newblock {\em arXiv preprint arXiv:1906.03351}, 2019.

\bibitem{banerjee2005meteor}
S.~Banerjee and A.~Lavie.
\newblock Meteor: An automatic metric for mt evaluation with improved
  correlation with human judgments.
\newblock In {\em Proceedings of the acl workshop on intrinsic and extrinsic
  evaluation measures for machine translation and/or summarization}, pages
  65--72, 2005.

\bibitem{barham2022pathways}
P.~Barham, A.~Chowdhery, J.~Dean, S.~Ghemawat, S.~Hand, D.~Hurt, M.~Isard,
  H.~Lim, R.~Pang, S.~Roy, et~al.
\newblock Pathways: Asynchronous distributed dataflow for ml.
\newblock {\em Proceedings of Machine Learning and Systems}, 4:430--449, 2022.

\bibitem{bevendorff2022overview}
J.~Bevendorff, B.~Chulvi, E.~Fersini, A.~Heini, M.~Kestemont, K.~Kredens,
  M.~Mayerl, R.~Ortega-Bueno, P.~P{\k{e}}zik, M.~Potthast, et~al.
\newblock Overview of pan 2022: Authorship verification, profiling irony and
  stereotype spreaders, style change detection, and trigger detection.
\newblock In {\em European Conference on Information Retrieval}, pages
  331--338. Springer, 2022.

\bibitem{bhat2020effectively}
M.~M. Bhat and S.~Parthasarathy.
\newblock How effectively can machines defend against machine-generated fake
  news? an empirical study.
\newblock In {\em Proceedings of the First Workshop on Insights from Negative
  Results in NLP}, pages 48--53, 2020.

\bibitem{biswas2023chatgpt}
S.~Biswas.
\newblock Chatgpt and the future of medical writing, 2023.

\bibitem{biten2019good}
A.~F. Biten, L.~Gomez, M.~Rusinol, and D.~Karatzas.
\newblock Good news, everyone! context driven entity-aware captioning for news
  images.
\newblock In {\em Proceedings of the IEEE/CVF Conference on Computer Vision and
  Pattern Recognition}, pages 12466--12475, 2019.

\bibitem{black2022gpt}
S.~Black, S.~Biderman, E.~Hallahan, Q.~Anthony, L.~Gao, L.~Golding, H.~He,
  C.~Leahy, K.~McDonell, J.~Phang, et~al.
\newblock Gpt-neox-20b: An open-source autoregressive language model.
\newblock In {\em Proceedings of BigScience Episode$\backslash$\# 5--Workshop
  on Challenges \& Perspectives in Creating Large Language Models}, pages
  95--136, 2022.

\bibitem{brown2020language}
T.~Brown, B.~Mann, N.~Ryder, M.~Subbiah, J.~D. Kaplan, P.~Dhariwal,
  A.~Neelakantan, P.~Shyam, G.~Sastry, A.~Askell, S.~Agarwal, A.~Herbert-Voss,
  G.~Krueger, T.~Henighan, R.~Child, A.~Ramesh, D.~Ziegler, J.~Wu, C.~Winter,
  C.~Hesse, M.~Chen, E.~Sigler, M.~Litwin, S.~Gray, B.~Chess, J.~Clark,
  C.~Berner, S.~McCandlish, A.~Radford, I.~Sutskever, and D.~Amodei.
\newblock Language models are few-shot learners.
\newblock In H.~Larochelle, M.~Ranzato, R.~Hadsell, M.~F. Balcan, and H.~Lin,
  editors, {\em Advances in Neural Information Processing Systems}, volume~33,
  pages 1877--1901. Curran Associates, Inc., 2020.

\bibitem{buchanan2021truth}
B.~Buchanan, A.~Lohn, M.~Musser, and K.~Sedova.
\newblock Truth, lies, and automation.
\newblock {\em Center for Security and Emerging Technology}, 2021.

\bibitem{cer2018universal}
D.~Cer, Y.~Yang, S.-y. Kong, N.~Hua, N.~Limtiaco, R.~S. John, N.~Constant,
  M.~Guajardo-Cespedes, S.~Yuan, C.~Tar, et~al.
\newblock Universal sentence encoder for english.
\newblock In {\em Proceedings of the 2018 conference on empirical methods in
  natural language processing: system demonstrations}, pages 169--174, 2018.

\bibitem{chen2020facebook}
P.-J. Chen, A.~Lee, C.~Wang, N.~Goyal, A.~Fan, M.~Williamson, and J.~Gu.
\newblock Facebook ai’s wmt20 news translation task submission.
\newblock In {\em Proceedings of the Fifth Conference on Machine Translation},
  pages 113--125, 2020.

\bibitem{chowdhery2022palm}
A.~Chowdhery, S.~Narang, J.~Devlin, M.~Bosma, G.~Mishra, A.~Roberts, P.~Barham,
  H.~W. Chung, C.~Sutton, S.~Gehrmann, et~al.
\newblock Palm: Scaling language modeling with pathways.
\newblock {\em arXiv preprint arXiv:2204.02311}, 2022.

\bibitem{clark2021all}
E.~Clark, T.~August, S.~Serrano, N.~Haduong, S.~Gururangan, and N.~A. Smith.
\newblock All that’s ‘human’is not gold: Evaluating human evaluation of
  generated text.
\newblock In {\em Proceedings of the 59th Annual Meeting of the Association for
  Computational Linguistics and the 11th International Joint Conference on
  Natural Language Processing (Volume 1: Long Papers)}, pages 7282--7296, 2021.

\bibitem{clark2020electra}
K.~Clark, M.-T. Luong, Q.~V. Le, and C.~D. Manning.
\newblock Electra: Pre-training text encoders as discriminators rather than
  generators.
\newblock {\em arXiv preprint arXiv:2003.10555}, 2020.

\bibitem{lample2019cross}
A.~Conneau and G.~Lample.
\newblock Cross-lingual language model pretraining.
\newblock {\em Advances in neural information processing systems}, 32, 2019.

\bibitem{cotton2023chatting}
D.~R. Cotton, P.~A. Cotton, and J.~R. Shipway.
\newblock Chatting and cheating. ensuring academic integrity in the era of
  chatgpt.
\newblock 2023.

\bibitem{crothers2022adversarial}
E.~Crothers, N.~Japkowicz, H.~Viktor, and P.~Branco.
\newblock Adversarial robustness of neural-statistical features in detection of
  generative transformers.
\newblock {\em arXiv preprint arXiv:2203.07983}, 2022.

\bibitem{cutlerautomatic}
J.~Cutler, L.~Dugan, S.~Havaldar, and A.~Stein.
\newblock Automatic detection of hybrid human-machine text boundaries.
\newblock 2021.

\bibitem{dai2019transformer}
Z.~Dai, Z.~Yang, Y.~Yang, J.~G. Carbonell, Q.~Le, and R.~Salakhutdinov.
\newblock Transformer-xl: Attentive language models beyond a fixed-length
  context.
\newblock In {\em Proceedings of the 57th Annual Meeting of the Association for
  Computational Linguistics}, pages 2978--2988, 2019.

\bibitem{dathathri2019plug}
S.~Dathathri, A.~Madotto, J.~Lan, J.~Hung, E.~Frank, P.~Molino, J.~Yosinski,
  and R.~Liu.
\newblock Plug and play language models: A simple approach to controlled text
  generation.
\newblock In {\em International Conference on Learning Representations}, 2020.

\bibitem{devlin2018bert}
J.~Devlin, M.-W. Chang, K.~Lee, and K.~Toutanova.
\newblock Bert: Pre-training of deep bidirectional transformers for language
  understanding.
\newblock {\em arXiv preprint arXiv:1810.04805}, 2018.

\bibitem{diwan2021fingerprinting}
N.~Diwan, T.~Chakraborty, and Z.~Shafiq.
\newblock Fingerprinting fine-tuned language models in the wild.
\newblock In {\em Findings of the Association for Computational Linguistics:
  ACL-IJCNLP 2021}, pages 4652--4664, 2021.

\bibitem{donahue2020enabling}
C.~Donahue, M.~Lee, and P.~Liang.
\newblock Enabling language models to fill in the blanks.
\newblock In {\em Proceedings of the 58th Annual Meeting of the Association for
  Computational Linguistics}, pages 2492--2501, 2020.

\bibitem{dou2021scarecrow}
Y.~Dou, M.~Forbes, R.~Koncel-Kedziorski, N.~A. Smith, and Y.~Choi.
\newblock Scarecrow: A framework for scrutinizing machine text.
\newblock {\em arXiv preprint arXiv:2107.01294}, 2021.

\bibitem{dowling2023chatgpt}
M.~Dowling and B.~Lucey.
\newblock Chatgpt for (finance) research: The bananarama conjecture.
\newblock {\em Finance Research Letters}, page 103662, 2023.

\bibitem{dugan2020roft}
L.~Dugan, D.~Ippolito, A.~Kirubarajan, and C.~Callison-Burch.
\newblock Roft: A tool for evaluating human detection of machine-generated
  text.
\newblock In {\em Proceedings of the 2020 Conference on Empirical Methods in
  Natural Language Processing: System Demonstrations}, pages 189--196, 2020.

\bibitem{fagni2021tweepfake}
T.~Fagni, F.~Falchi, M.~Gambini, A.~Martella, and M.~Tesconi.
\newblock Tweepfake: About detecting deepfake tweets.
\newblock {\em Plos one}, 16(5):e0251415, 2021.

\bibitem{fan2018hierarchical}
A.~Fan, M.~Lewis, and Y.~Dauphin.
\newblock Hierarchical neural story generation.
\newblock In {\em Proceedings of the 56th Annual Meeting of the Association for
  Computational Linguistics (Volume 1: Long Papers)}, pages 889--898, 2018.

\bibitem{fast2016empath}
E.~Fast, B.~Chen, and M.~S. Bernstein.
\newblock Empath: Understanding topic signals in large-scale text.
\newblock In {\em Proceedings of the 2016 CHI conference on human factors in
  computing systems}, pages 4647--4657, 2016.

\bibitem{fedus2022switch}
W.~Fedus, B.~Zoph, and N.~Shazeer.
\newblock Switch transformers: Scaling to trillion parameter models with simple
  and efficient sparsity.
\newblock {\em Journal of Machine Learning Research}, 23(120):1--39, 2022.

\bibitem{frohling2021feature}
L.~Fr{\"o}hling and A.~Zubiaga.
\newblock Feature-based detection of automated language models: tackling gpt-2,
  gpt-3 and grover.
\newblock {\em PeerJ Computer Science}, 7:e443, 2021.

\bibitem{gagiano2021robustness}
R.~Gagiano, M.~M.-H. Kim, X.~J. Zhang, and J.~Biggs.
\newblock Robustness analysis of grover for machine-generated news detection.
\newblock In {\em Proceedings of the The 19th Annual Workshop of the
  Australasian Language Technology Association}, pages 119--127, 2021.

\bibitem{galle2021unsupervised}
M.~Gall{\'e}, J.~Rozen, G.~Kruszewski, and H.~Elsahar.
\newblock Unsupervised and distributional detection of machine-generated text.
\newblock {\em arXiv preprint arXiv:2111.02878}, 2021.

\bibitem{gambini2022pushing}
M.~Gambini, T.~Fagni, F.~Falchi, and M.~Tesconi.
\newblock On pushing deepfake tweet detection capabilities to the limits.
\newblock In {\em 14th ACM Web Science Conference 2022}, pages 154--163, 2022.

\bibitem{gao2022comparing}
C.~A. Gao, F.~M. Howard, N.~S. Markov, E.~C. Dyer, S.~Ramesh, Y.~Luo, and A.~T.
  Pearson.
\newblock Comparing scientific abstracts generated by chatgpt to original
  abstracts using an artificial intelligence output detector, plagiarism
  detector, and blinded human reviewers.
\newblock {\em bioRxiv}, pages 2022--12, 2022.

\bibitem{gao2018black}
J.~Gao, J.~Lanchantin, M.~L. Soffa, and Y.~Qi.
\newblock Black-box generation of adversarial text sequences to evade deep
  learning classifiers.
\newblock In {\em 2018 IEEE Security and Privacy Workshops (SPW)}, pages
  50--56. IEEE, 2018.

\bibitem{gao2020pile}
L.~Gao, S.~Biderman, S.~Black, L.~Golding, T.~Hoppe, C.~Foster, J.~Phang,
  H.~He, A.~Thite, N.~Nabeshima, et~al.
\newblock The pile: An 800gb dataset of diverse text for language modeling.
\newblock {\em arXiv preprint arXiv:2101.00027}, 2020.

\bibitem{gao2021simcse}
T.~Gao, X.~Yao, and D.~Chen.
\newblock Simcse: Simple contrastive learning of sentence embeddings.
\newblock In {\em Proceedings of the 2021 Conference on Empirical Methods in
  Natural Language Processing}, pages 6894--6910, 2021.

\bibitem{gehman2020realtoxicityprompts}
S.~Gehman, S.~Gururangan, M.~Sap, Y.~Choi, and N.~A. Smith.
\newblock Realtoxicityprompts: Evaluating neural toxic degeneration in language
  models.
\newblock In {\em Findings of the Association for Computational Linguistics:
  EMNLP 2020}, pages 3356--3369, 2020.

\bibitem{gehrmann2019gltr}
S.~Gehrmann, H.~Strobelt, and A.~M. Rush.
\newblock Gltr: Statistical detection and visualization of generated text.
\newblock In {\em Proceedings of the 57th Annual Meeting of the Association for
  Computational Linguistics: System Demonstrations}, pages 111--116, 2019.

\bibitem{guerrero2022synthetic}
J.~Guerrero and I.~Alsmadi.
\newblock Synthetic text detection: Systemic literature review.
\newblock {\em arXiv preprint arXiv:2210.06336}, 2022.

\bibitem{haroon2021avengers}
M.~Haroon, F.~Zaffar, P.~Srinivasan, and Z.~Shafiq.
\newblock Avengers ensemble! improving transferability of authorship
  obfuscation.
\newblock {\em arXiv preprint arXiv:2109.07028}, 2021.

\bibitem{he2021debertav3}
P.~He, J.~Gao, and W.~Chen.
\newblock Debertav3: Improving deberta using electra-style pre-training with
  gradient-disentangled embedding sharing.
\newblock {\em arXiv preprint arXiv:2111.09543}, 2021.

\bibitem{he2021deberta}
P.~He, X.~Liu, J.~Gao, and W.~Chen.
\newblock Deberta: Decoding-enhanced bert with disentangled attention.
\newblock In {\em International Conference on Learning Representations}, 2021.

\bibitem{holtzman2019curious}
A.~Holtzman, J.~Buys, L.~Du, M.~Forbes, and Y.~Choi.
\newblock The curious case of neural text degeneration.
\newblock In {\em International Conference on Learning Representations}, 2019.

\bibitem{gpt2outputdetector}
Huggingface.
\newblock Gpt-2 output detector demo.
\newblock {\em https://huggingface.co/openai-detector/}, 2019.

\bibitem{ippolito2020automatic}
D.~Ippolito, D.~Duckworth, C.~Callison-Burch, and D.~Eck.
\newblock Automatic detection of generated text is easiest when humans are
  fooled.
\newblock In {\em Proceedings of the 58th Annual Meeting of the Association for
  Computational Linguistics}, pages 1808--1822, 2020.

\bibitem{jafariakinabad2019syntactic}
F.~Jafariakinabad, S.~Tarnpradab, and K.~A. Hua.
\newblock Syntactic recurrent neural network for authorship attribution.
\newblock {\em arXiv preprint arXiv:1902.09723}, 2019.

\bibitem{jawahar2022automatic}
G.~Jawahar, M.~Abdul-Mageed, and L.~Lakshmanan.
\newblock Automatic detection of entity-manipulated text using factual
  knowledge.
\newblock In {\em Proceedings of the 60th Annual Meeting of the Association for
  Computational Linguistics (Volume 2: Short Papers)}, pages 86--93, 2022.

\bibitem{jin2020bert}
D.~Jin, Z.~Jin, J.~T. Zhou, and P.~Szolovits.
\newblock Is bert really robust? a strong baseline for natural language attack
  on text classification and entailment.
\newblock In {\em Proceedings of the AAAI conference on artificial
  intelligence}, volume~34, pages 8018--8025, 2020.

\bibitem{jones2022you}
K.~Jones, J.~R. Nurse, and S.~Li.
\newblock Are you robert or roberta? deceiving online authorship attribution
  models using neural text generators.
\newblock In {\em Proceedings of the International AAAI Conference on Web and
  Social Media}, volume~16, pages 429--440, 2022.

\bibitem{karadzhov2017case}
G.~Karadzhov, T.~Mihaylova, Y.~Kiprov, G.~Georgiev, I.~Koychev, and P.~Nakov.
\newblock The case for being average: A mediocrity approach to style masking
  and author obfuscation.
\newblock In {\em International Conference of the Cross-Language Evaluation
  Forum for European Languages}, pages 173--185. Springer, 2017.

\bibitem{kaur2019authorship}
R.~Kaur, S.~Singh, and H.~Kumar.
\newblock Authorship analysis of online social media content.
\newblock In {\em Proceedings of 2nd International Conference on Communication,
  Computing and Networking}, pages 539--549. Springer, 2019.

\bibitem{keskar2019ctrl}
N.~S. Keskar, B.~McCann, L.~R. Varshney, C.~Xiong, and R.~Socher.
\newblock Ctrl: A conditional transformer language model for controllable
  generation.
\newblock {\em arXiv preprint arXiv:1909.05858}, 2019.

\bibitem{kestemont2021overview}
M.~Kestemont, E.~Manjavacas, I.~Markov, J.~Bevendorff, M.~Wiegmann,
  E.~Stamatatos, B.~Stein, and M.~Potthast.
\newblock Overview of the cross-domain authorship verification task at pan
  2021.
\newblock In {\em CLEF (Working Notes)}, 2021.

\bibitem{kipf2016semi}
T.~N. Kipf and M.~Welling.
\newblock Semi-supervised classification with graph convolutional networks.
\newblock {\em arXiv preprint arXiv:1609.02907}, 2016.

\bibitem{kirchenbauer2023watermark}
J.~Kirchenbauer, J.~Geiping, Y.~Wen, J.~Katz, I.~Miers, and T.~Goldstein.
\newblock A watermark for large language models.
\newblock {\em arXiv preprint arXiv:2301.10226}, 2023.

\bibitem{krause2021gedi}
B.~Krause, A.~D. Gotmare, B.~McCann, N.~S. Keskar, S.~Joty, R.~Socher, and
  N.~F. Rajani.
\newblock Gedi: Generative discriminator guided sequence generation.
\newblock In {\em Findings of the Association for Computational Linguistics:
  EMNLP 2021}, pages 4929--4952, 2021.

\bibitem{krause2016multiplicative}
B.~Krause, L.~Lu, I.~Murray, and S.~Renals.
\newblock Multiplicative lstm for sequence modelling.
\newblock {\em arXiv preprint arXiv:1609.07959}, 2016.

\bibitem{krause2017multiplicative}
B.~Krause, I.~Murray, S.~Renals, and L.~Liang.
\newblock Multiplicative lstm for sequence modelling.
\newblock In {\em 5th International Conference on Learning Representations},
  pages 2872--2880, 2017.

\bibitem{kreps2022all}
S.~Kreps, R.~M. McCain, and M.~Brundage.
\newblock All the news that’s fit to fabricate: Ai-generated text as a tool
  of media misinformation.
\newblock {\em Journal of Experimental Political Science}, 9(1):104--117, 2022.

\bibitem{kushnareva2021artificial}
L.~Kushnareva, D.~Cherniavskii, V.~Mikhailov, E.~Artemova, S.~Barannikov,
  A.~Bernstein, I.~Piontkovskaya, D.~Piontkovski, and E.~Burnaev.
\newblock Artificial text detection via examining the topology of attention
  maps.
\newblock In {\em Proceedings of the 2021 Conference on Empirical Methods in
  Natural Language Processing}, pages 635--649, 2021.

\bibitem{lagutina2019survey}
K.~Lagutina, N.~Lagutina, E.~Boychuk, I.~Vorontsova, E.~Shliakhtina,
  O.~Belyaeva, I.~Paramonov, and P.~Demidov.
\newblock A survey on stylometric text features.
\newblock In {\em 2019 25th Conference of Open Innovations Association
  (FRUCT)}, pages 184--195. IEEE, 2019.

\bibitem{lan2019albert}
Z.~Lan, M.~Chen, S.~Goodman, K.~Gimpel, P.~Sharma, and R.~Soricut.
\newblock Albert: A lite bert for self-supervised learning of language
  representations.
\newblock {\em arXiv preprint arXiv:1909.11942}, 2019.

\bibitem{lecun2006tutorial}
Y.~LeCun, S.~Chopra, R.~Hadsell, M.~Ranzato, and F.~Huang.
\newblock A tutorial on energy-based learning.
\newblock {\em Predicting structured data}, 1(0), 2006.

\bibitem{lewis2020bart}
M.~Lewis, Y.~Liu, N.~Goyal, M.~Ghazvininejad, A.~Mohamed, O.~Levy, V.~Stoyanov,
  and L.~Zettlemoyer.
\newblock Bart: Denoising sequence-to-sequence pre-training for natural
  language generation, translation, and comprehension.
\newblock In {\em Proceedings of the 58th Annual Meeting of the Association for
  Computational Linguistics}, pages 7871--7880, 2020.

\bibitem{li2021pretrained}
J.~Li, T.~Tang, W.~X. Zhao, and J.-R. Wen.
\newblock Pretrained language models for text generation: A survey.
\newblock {\em arXiv preprint arXiv:2105.10311}, 2021.

\bibitem{liu2022coco}
X.~Liu, Z.~Zhang, Y.~Wang, Y.~Lan, and C.~Shen.
\newblock Coco: Coherence-enhanced machine-generated text detection under data
  limitation with contrastive learning.
\newblock {\em arXiv preprint arXiv:2212.10341}, 2022.

\bibitem{liu2019roberta}
Y.~Liu, M.~Ott, N.~Goyal, J.~Du, M.~Joshi, D.~Chen, O.~Levy, M.~Lewis,
  L.~Zettlemoyer, and V.~Stoyanov.
\newblock Roberta: A robustly optimized bert pretraining approach.
\newblock {\em arXiv preprint arXiv:1907.11692}, 2019.

\bibitem{liyanage2022benchmark}
V.~Liyanage, D.~Buscaldi, and A.~Nazarenko.
\newblock A benchmark corpus for the detection of automatically generated text
  in academic publications.
\newblock In {\em LREC}, 2022.

\bibitem{mahmood2019girl}
A.~Mahmood, F.~Ahmad, Z.~Shafiq, P.~Srinivasan, and F.~Zaffar.
\newblock A girl has no name: Automated authorship obfuscation using mutant-x.
\newblock {\em Proc. Priv. Enhancing Technol.}, 2019(4):54--71, 2019.

\bibitem{mahmood2020girl}
A.~Mahmood, Z.~Shafiq, and P.~Srinivasan.
\newblock A girl has a name: Detecting authorship obfuscation.
\newblock In {\em Proceedings of the 58th Annual Meeting of the Association for
  Computational Linguistics}, pages 2235--2245, 2020.

\bibitem{mcdonald2012use}
A.~W. McDonald, S.~Afroz, A.~Caliskan, A.~Stolerman, and R.~Greenstadt.
\newblock Use fewer instances of the letter “i”: Toward writing style
  anonymization.
\newblock In {\em International Symposium on Privacy Enhancing Technologies
  Symposium}, pages 299--318. Springer, 2012.

\bibitem{mirsky2021creation}
Y.~Mirsky and W.~Lee.
\newblock The creation and detection of deepfakes: A survey.
\newblock {\em ACM Computing Surveys (CSUR)}, 54(1):1--41, 2021.

\bibitem{mitchell2023detectgpt}
E.~Mitchell, Y.~Lee, A.~Khazatsky, C.~D. Manning, and C.~Finn.
\newblock Detectgpt: Zero-shot machine-generated text detection using
  probability curvature.
\newblock {\em arXiv preprint arXiv:2301.11305}, 2023.

\bibitem{munir2021through}
S.~Munir, B.~Batool, Z.~Shafiq, P.~Srinivasan, and F.~Zaffar.
\newblock Through the looking glass: Learning to attribute synthetic text
  generated by language models.
\newblock In {\em Proceedings of the 16th Conference of the European Chapter of
  the Association for Computational Linguistics: Main Volume}, pages
  1811--1822, 2021.

\bibitem{ng2019facebook}
N.~Ng, K.~Yee, A.~Baevski, M.~Ott, M.~Auli, and S.~Edunov.
\newblock Facebook fair's wmt19 news translation task submission.
\newblock {\em arXiv preprint arXiv:1907.06616}, 2019.

\bibitem{nguyen2017identifying}
H.-Q. Nguyen-Son, N.-D.~T. Tieu, H.~H. Nguyen, J.~Yamagishi, and I.~E. Zen.
\newblock Identifying computer-generated text using statistical analysis.
\newblock In {\em 2017 Asia-Pacific Signal and Information Processing
  Association Annual Summit and Conference (APSIPA ASC)}, pages 1504--1511.
  IEEE, 2017.

\bibitem{chatgpt}
OpenAI.
\newblock Optimizing language models for dialogue.
\newblock {\em https://openai.com/blog/chatgpt/}, 2022.

\bibitem{automik22}
M.~Orzhenovskii.
\newblock Detecting auto-generated texts with language model and attacking the
  detector.
\newblock {\em Computational Linguistics and Intellectual Technologies:
  Proceedings of the International Conference “Dialogue 2022}, 2022.

\bibitem{pagnoni2022threat}
A.~Pagnoni, M.~Graciarena, and Y.~Tsvetkov.
\newblock Threat scenarios and best practices to detect neural fake news.
\newblock In {\em Proceedings of the 29th International Conference on
  Computational Linguistics}, pages 1233--1249, 2022.

\bibitem{pennebaker2001linguistic}
J.~W. Pennebaker, M.~E. Francis, and R.~J. Booth.
\newblock Linguistic inquiry and word count: Liwc 2001.
\newblock {\em Mahway: Lawrence Erlbaum Associates}, 71(2001):2001, 2001.

\bibitem{pennington2014glove}
J.~Pennington, R.~Socher, and C.~D. Manning.
\newblock Glove: Global vectors for word representation.
\newblock In {\em Proceedings of the 2014 conference on empirical methods in
  natural language processing (EMNLP)}, pages 1532--1543, 2014.

\bibitem{pillutla2021information}
K.~Pillutla, S.~Swayamdipta, R.~Zellers, J.~Thickstun, S.~Welleck, Y.~Choi, and
  Z.~Harchaoui.
\newblock An information divergence measure between neural text and human text.
\newblock {\em arXiv preprint arXiv:2102.01454}, 2021.

\bibitem{puunraveling}
J.~Pu, Z.~Huang, Y.~Xi, G.~Chen, W.~Chen, and R.~Zhang.
\newblock Unraveling the mystery of artifacts in machine generated text.
\newblock pages s 6889--6898, 2022.

\bibitem{pudeepfake}
J.~Pu, Z.~Sarwar, S.~M. Abdullah, A.~Rehman, Y.~Kim, P.~Bhattacharya, M.~Javed,
  B.~Viswanath, V.~Tech, and L.~Pakistan.
\newblock Deepfake text detection: Limitations and opportunities.
\newblock {\em 44th IEEE Symposium on Security and Privacy}, 2023.

\bibitem{radford2018improving}
A.~Radford, K.~Narasimhan, T.~Salimans, and I.~Sutskever.
\newblock Improving language understanding by generative pre-training.
\newblock {\em URL https://s3-us-west-2. amazonaws.
  com/openai-assets/researchcovers/languageunsupervised/language understanding
  paper. pdf}, 2018.

\bibitem{radford2019language}
A.~Radford, J.~Wu, R.~Child, D.~Luan, D.~Amodei, and I.~Sutskever.
\newblock Language models are unsupervised multitask learners.
\newblock {\em OpenAI blog}, 1(8):9, 2019.

\bibitem{raffel2020exploring}
C.~Raffel, N.~Shazeer, A.~Roberts, K.~Lee, S.~Narang, M.~Matena, Y.~Zhou,
  W.~Li, P.~J. Liu, et~al.
\newblock Exploring the limits of transfer learning with a unified text-to-text
  transformer.
\newblock {\em J. Mach. Learn. Res.}, 21(140):1--67, 2020.

\bibitem{rosati2022synscipass}
D.~Rosati.
\newblock Synscipass: detecting appropriate uses of scientific text generation.
\newblock In {\em Proceedings of the Third Workshop on Scholarly Document
  Processing}, pages 214--222, 2022.

\bibitem{schuster2020limitations}
T.~Schuster, R.~Schuster, D.~J. Shah, and R.~Barzilay.
\newblock The limitations of stylometry for detecting machine-generated fake
  news.
\newblock {\em Computational Linguistics}, 46(2):499--510, 2020.

\bibitem{shamardina2022findings}
T.~Shamardina, V.~Mikhailov, D.~Chernianskii, A.~Fenogenova, M.~Saidov,
  A.~Valeeva, T.~Shavrina, I.~Smurov, E.~Tutubalina, and E.~Artemova.
\newblock Findings of the the ruatd shared task 2022 on artificial text
  detection in russian.
\newblock {\em arXiv preprint arXiv:2206.01583}, 2022.

\bibitem{shao2019reverse}
J.~Shao, A.~Uchendu, and D.~Lee.
\newblock A reverse turing test for detecting machine-made texts.
\newblock In {\em Proceedings of the 10th ACM Conference on Web Science}, pages
  275--279, 2019.

\bibitem{shrestha2017convolutional}
P.~Shrestha, S.~Sierra, F.~Gonzalez, M.~Montes, P.~Rosso, and T.~Solorio.
\newblock Convolutional neural networks for authorship attribution of short
  texts.
\newblock In {\em Proceedings of the 15th Conference of the European Chapter of
  the Association for Computational Linguistics: Volume 2, Short Papers}, 2017.

\bibitem{stamatatos2016authorship}
E.~Stamatatos.
\newblock Authorship verification: a review of recent advances.
\newblock {\em Research in Computing Science}, 123:9--25, 2016.

\bibitem{stamatatos2022overview}
E.~Stamatatos, M.~Kestemont, K.~Kredens, P.~Pezik, A.~Heini, J.~Bevendorff,
  M.~Potthast, and B.~Stein.
\newblock Overview of the authorship verification task at pan 2022.
\newblock {\em Working Notes of CLEF}, 2022.

\bibitem{stiff2022detecting}
H.~Stiff and F.~Johansson.
\newblock Detecting computer-generated disinformation.
\newblock {\em International Journal of Data Science and Analytics},
  13(4):363--383, 2022.

\bibitem{tabatabaei2015survey}
M.~Tabatabaei, J.~Hakanen, M.~Hartikainen, K.~Miettinen, and K.~Sindhya.
\newblock A survey on handling computationally expensive multiobjective
  optimization problems using surrogates: non-nature inspired methods.
\newblock {\em Structural and Multidisciplinary Optimization}, 52(1):1--25,
  2015.

\bibitem{tan2020detecting}
R.~Tan, B.~Plummer, and K.~Saenko.
\newblock Detecting cross-modal inconsistency to defend against neural fake
  news.
\newblock In {\em Proceedings of the 2020 Conference on Empirical Methods in
  Natural Language Processing (EMNLP)}, pages 2081--2106, 2020.

\bibitem{tyo2022state}
J.~Tyo, B.~Dhingra, and Z.~C. Lipton.
\newblock On the state of the art in authorship attribution and authorship
  verification.
\newblock {\em arXiv preprint arXiv:2209.06869}, 2022.

\bibitem{uchendu2019characterizing}
A.~Uchendu, J.~Cao, Q.~Wang, B.~Luo, and D.~Lee.
\newblock Characterizing man-made vs. machine-made chatbot dialogs.
\newblock In {\em TTO}, 2019.

\bibitem{uchendu2020authorship}
A.~Uchendu, T.~Le, K.~Shu, and D.~Lee.
\newblock Authorship attribution for neural text generation.
\newblock In {\em Proceedings of the 2020 Conference on Empirical Methods in
  Natural Language Processing (EMNLP)}, pages 8384--8395, 2020.

\bibitem{uchendu2021turingbench}
A.~Uchendu, Z.~Ma, T.~Le, R.~Zhang, and D.~Lee.
\newblock Turingbench: A benchmark environment for turing test in the age of
  neural text generation.
\newblock In {\em Findings of the Association for Computational Linguistics:
  EMNLP 2021}, pages 2001--2016, 2021.

\bibitem{varol2017online}
O.~Varol, E.~Ferrara, C.~Davis, F.~Menczer, and A.~Flammini.
\newblock Online human-bot interactions: Detection, estimation, and
  characterization.
\newblock In {\em Proceedings of the international AAAI conference on web and
  social media}, volume~11, pages 280--289, 2017.

\bibitem{varshney2020limits}
L.~R. Varshney, N.~S. Keskar, and R.~Socher.
\newblock Limits of detecting text generated by large-scale language models.
\newblock In {\em 2020 Information Theory and Applications Workshop (ITA)},
  pages 1--5. IEEE, 2020.

\bibitem{meshtransformerjax}
B.~Wang.
\newblock {Mesh-Transformer-JAX: Model-Parallel Implementation of Transformer
  Language Model with JAX}.
\newblock \url{https://github.com/kingoflolz/mesh-transformer-jax}, May 2021.

\bibitem{gpt-j}
B.~Wang and A.~Komatsuzaki.
\newblock {GPT-J-6B: A 6 Billion Parameter Autoregressive Language Model}.
\newblock \url{https://github.com/kingoflolz/mesh-transformer-jax}, May 2021.

\bibitem{wolf2019huggingface}
T.~Wolf, L.~Debut, V.~Sanh, J.~Chaumond, C.~Delangue, A.~Moi, P.~Cistac,
  T.~Rault, R.~Louf, M.~Funtowicz, et~al.
\newblock Huggingface's transformers: State-of-the-art natural language
  processing.
\newblock {\em arXiv preprint arXiv:1910.03771}, 2019.

\bibitem{wolff2020attacking}
M.~Wolff and S.~Wolff.
\newblock Attacking neural text detectors.
\newblock {\em arXiv preprint arXiv:2002.11768}, 2020.

\bibitem{alison}
E.~Xing, T.~Le, and D.~Lee.
\newblock Alison: Fast stylometric authorship obfuscation.
\newblock {\em Technical Report}, 2023.

\bibitem{yang2019xlnet}
Z.~Yang, Z.~Dai, Y.~Yang, J.~Carbonell, R.~R. Salakhutdinov, and Q.~V. Le.
\newblock Xlnet: Generalized autoregressive pretraining for language
  understanding.
\newblock In {\em Advances in neural information processing systems}, pages
  5754--5764, 2019.

\bibitem{zellers2019defending}
R.~Zellers, A.~Holtzman, H.~Rashkin, Y.~Bisk, A.~Farhadi, F.~Roesner, and
  Y.~Choi.
\newblock Defending against neural fake news.
\newblock In {\em Advances in Neural Information Processing Systems}, pages
  9051--9062, 2019.

\bibitem{zhai2022girl}
W.~Zhai, J.~Rusert, Z.~Shafiq, and P.~Srinivasan.
\newblock A girl has a name, and it's... adversarial authorship attribution for
  deobfuscation.
\newblock {\em arXiv preprint arXiv:2203.11849}, 2022.

\bibitem{zhang2022survey}
H.~Zhang, H.~Song, S.~Li, M.~Zhou, and D.~Song.
\newblock A survey of controllable text generation using transformer-based
  pre-trained language models.
\newblock {\em arXiv preprint arXiv:2201.05337}, 2022.

\bibitem{zhang2018syntax}
R.~Zhang, Z.~Hu, H.~Guo, and Y.~Mao.
\newblock Syntax encoding with application in authorship attribution.
\newblock In {\em Proceedings of the 2018 Conference on Empirical Methods in
  Natural Language Processing}, pages 2742--2753, 2018.

\bibitem{zhang2022opt}
S.~Zhang, S.~Roller, N.~Goyal, M.~Artetxe, M.~Chen, S.~Chen, C.~Dewan, M.~Diab,
  X.~Li, X.~V. Lin, et~al.
\newblock Opt: Open pre-trained transformer language models.
\newblock {\em arXiv preprint arXiv:2205.01068}, 2022.

\bibitem{zhang2020adversarial}
W.~E. Zhang, Q.~Z. Sheng, A.~Alhazmi, and C.~Li.
\newblock Adversarial attacks on deep-learning models in natural language
  processing: A survey.
\newblock {\em ACM Transactions on Intelligent Systems and Technology (TIST)},
  11(3):1--41, 2020.

\bibitem{zhong2020neural}
W.~Zhong, D.~Tang, Z.~Xu, R.~Wang, N.~Duan, M.~Zhou, J.~Wang, and J.~Yin.
\newblock Neural deepfake detection with factual structure of text.
\newblock In {\em Proceedings of the 2020 Conference on Empirical Methods in
  Natural Language Processing (EMNLP)}, pages 2461--2470, 2020.

\bibitem{zhu2020gruen}
W.~Zhu and S.~Bhat.
\newblock Gruen for evaluating linguistic quality of generated text.
\newblock In {\em Findings of the Association for Computational Linguistics:
  EMNLP 2020}, pages 94--108, 2020.

\end{thebibliography}

% \appendix

\end{document}